\pdfoutput=1

\documentclass[11pt]{article}

\usepackage[final]{acl}

\usepackage{times}
\usepackage{latexsym}

\usepackage[T1]{fontenc}

\usepackage[utf8]{inputenc}

\usepackage{microtype}

\usepackage{inconsolata}

\usepackage{graphicx}

\title{Benchmarking Deflection and Hallucination \\ in Large Vision-Language Models} 

\author{
 \textbf{Nicholas Moratelli\textsuperscript{$\dagger$}\thanks{Work done during an internship at Amazon.}},
 \textbf{Christopher Davis\textsuperscript{$\ddagger$}},
 \textbf{Leonardo F. R. Ribeiro\textsuperscript{$\ddagger$}},
 \textbf{Bill Byrne\textsuperscript{$\ddagger$,$\mathsection$}},
\\
 \textbf{Gonzalo Iglesias\textsuperscript{$\ddagger$}},
\\
 \textsuperscript{$\dagger$}University of Modena and Reggio Emilia ,
 \textsuperscript{$\ddagger$}Amazon AGI,
 \textsuperscript{$\mathsection$}University of Cambridge
\\
 \small{
   \textbf{Correspondence:} {davisjnh@amazon.com}
 }
}

\usepackage{booktabs}
\usepackage{xcolor}
\usepackage{colortbl}
\usepackage{booktabs}
\usepackage{amssymb}
\usepackage{amsmath}
\usepackage{multirow}
\usepackage{array}
\usepackage{makecell}
\usepackage{graphicx}
\usepackage{tabularx}
\usepackage[most]{tcolorbox}
\usepackage[ruled,vlined]{algorithm2e}
\usepackage{float}

\definecolor{headercolor}{RGB}{240, 240, 240}
\definecolor{ourlightgray}{RGB}{240, 240, 240}
\definecolor{CorrectColor}{RGB}{34,139,34}
\definecolor{DeflectColor}{RGB}{255,140,0}
\definecolor{HallucinateColor}{RGB}{178,34,34}

\newcommand{\judge}[2]{\makebox[\linewidth][r]{\textsc{Judge: \textcolor{#1}{#2}}}}

\newcommand{\kw}[1]{\textcolor{blue!70!black}{\textbf{#1}}}
\newcommand{\var}[1]{\textcolor{teal!80!black}{#1}}
\newcommand{\const}[1]{%
  \textcolor{orange!80!black}{\ensuremath{\text{\ttfamily\upshape #1}}}%
}

\newcommand{\cmark}{\textcolor{green!70!black}{$\checkmark$}}
\newcommand{\xmark}{\textcolor{red!70!black}{$\times$}}
\newcommand{\ours}{\textbf{VLM-DeflectionBench}} 

\begin{document}
\maketitle
\begin{abstract}
Large Vision–Language Models (LVLMs) increasingly rely on retrieval to answer knowledge-intensive \textit{multimodal} questions. Existing benchmarks overlook conflicts between visual and textual evidence and the importance of generating \textit{deflections} (\textit{e.g.}, ``Sorry, I cannot answer...'') when retrieved knowledge is incomplete. These benchmarks also suffer from rapid obsolescence, as growing LVLM training sets allow models to answer many questions without retrieval. We address these gaps with three contributions. First, we propose a dynamic data curation pipeline that preserves benchmark difficulty over time by filtering for genuinely retrieval-dependent samples. Second, we introduce \ours, a benchmark of $2,775$ samples spanning diverse multimodal retrieval settings, designed to probe model behaviour under conflicting or insufficient evidence. Third, we define a fine-grained evaluation protocol with four scenarios that disentangle parametric memorization from retrieval robustness. Experiments across 20 state-of-the-art LVLMs indicate that models usually fail to deflect in the presence of noisy or misleading evidence. Our results highlight the need to evaluate not only what models know, but how they behave when they do not, and serve as a reusable and extensible benchmark for reliable KB-VQA evaluation. All resources will be publicly available upon publication.
\end{abstract}

\begin{figure}[!ht]
  \includegraphics[width=\linewidth]{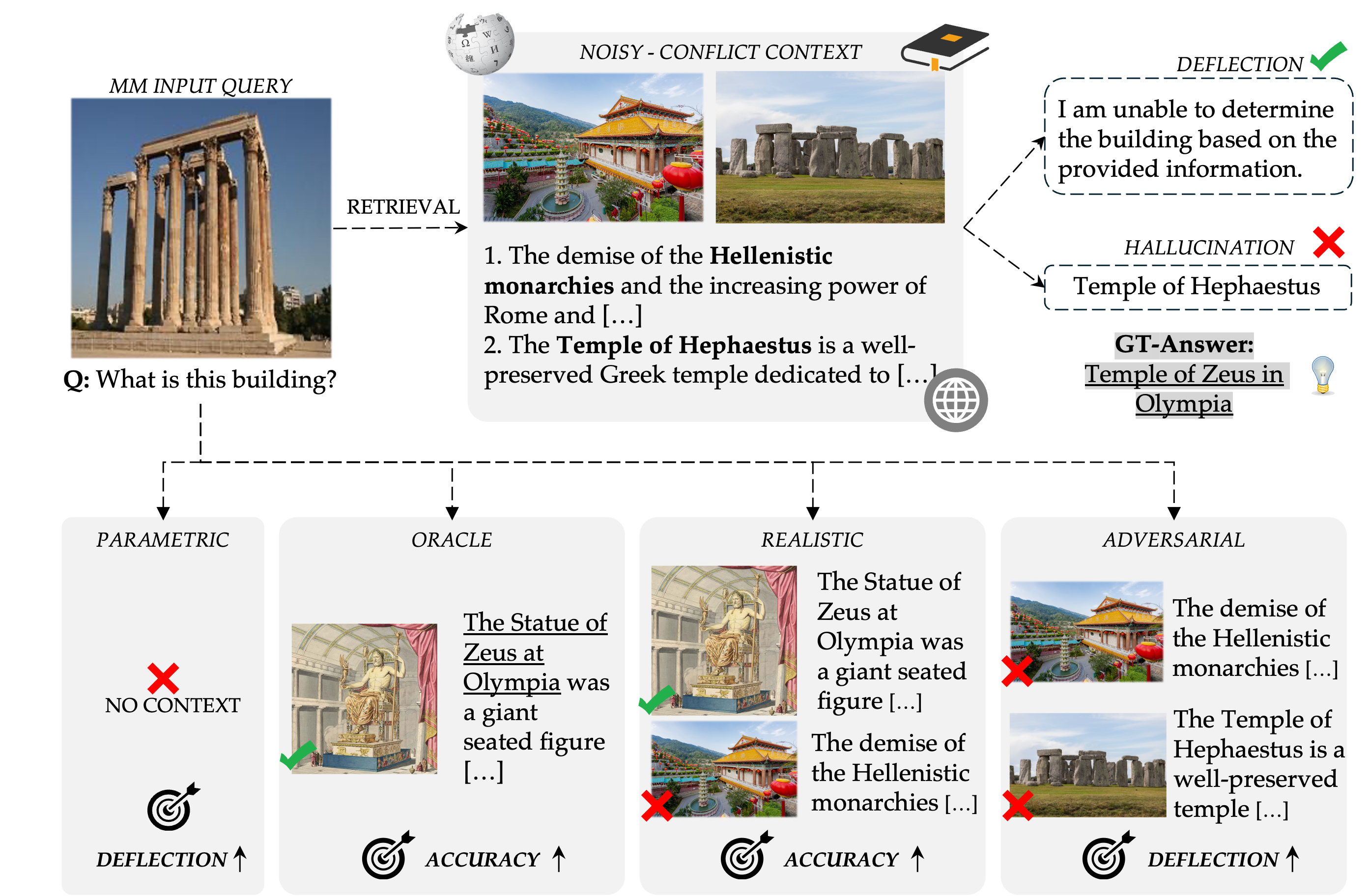}
  \caption{Overview of \ours. Top: LVLMs often hallucinate instead of abstaining when context is misleading. Bottom: \ours\ evaluates calibration across four scenarios-\textbf{Parametric}, \textbf{Oracle}, \textbf{Realistic}, and \textbf{Adversarial} to test whether models align their behavior with available knowledge.}
  \label{fig:main_figure}
\end{figure}

\section{Introduction}
Large Vision–Language Models (LVLMs) are rapidly moving into real-world applications where reliability is critical~\cite{caffagni2024revolution, liu2024improved}. Users expect not only accurate answers to complex multimodal queries, but also trustful behavior when knowledge is missing or contradictory. Indeed, LVLMs prompted to strictly ground their answers in retrieved evidence should state that the user request cannot be fulfilled when said knowledge does not support a reliable answer. Instead, models might generate an answer that cannot be traced back to correctly retrieved evidence. In this paper, we will refer to these two types of answer as \textbf{deflection} and \textbf{hallucination}, respectively.  We note that both types of answers are strictly speaking incorrect in that the user request is not fulfilled. However, when faced with insufficient evidence, deflection is the preferable failure mode.

Knowledge-based Visual Question Answering (KB-VQA) provides a natural testbed to study these behaviors. In KB-VQA, models must integrate both visual inputs and retrieved textual or visual passages to answer open-ended questions. Several benchmarks~\cite{mensink2023encyclopedic, chen2023can, chang2022webqa, lerner2022viquae, humrag} have addressed these problems, but they face two limitations. First, they suffer from \textit{rapid obsolescence}: many questions that once required retrieval can now be answered directly from parametric knowledge. Second, they focus on \textit{accuracy}, but do not distinguish failure modes.

To close this gap we present \textbf{VLM-DeflectionBench}, a new benchmark with a strict retrieval-augmented generation evaluation. As illustrated in  Figure~\ref{fig:main_figure}, models are judged not only on what they know, but also on how they behave when knowledge is missing or misleading. An ideal system should abstain when reliable evidence is absent, rather than hallucinate unsupported answers.

\begin{table*}[!ht]
\centering
\scriptsize
\setlength{\tabcolsep}{4pt}
\begin{tabular}{lcccccccc}
\toprule
\multirow{2}{*}[-0.6ex]{Benchmark} & \multicolumn{4}{c}{\textbf{Evaluation Scope}} & \multicolumn{4}{c}{\textbf{Evidence Configuration}} \\
\cmidrule(lr){2-5} \cmidrule(lr){6-9}
 & Hallucination & Deflection & Scenarios & Dynamic & MM-Query & MM-Context & Pre-retrieved & Negatives \\
\midrule
OK-VQA~\cite{marino2019ok}     & \xmark & \xmark & 1 & \xmark & \cmark & \xmark & \xmark & \xmark \\
A-OKVQA~\cite{schwenk2022okvqa}    & \xmark & \xmark & 1 & \xmark & \cmark & \xmark & \xmark & \xmark \\
E-VQA~\cite{mensink2023encyclopedic}      & \xmark & \xmark & 1 & \xmark & \cmark & \xmark & \xmark & \xmark \\
InfoSeek~\cite{chen2023can}   & \xmark & \xmark & 1 & \xmark & \cmark & \xmark & \xmark & \xmark \\
WebQA~\cite{chang2022webqa}      & \xmark & \xmark & 1 & \xmark & \xmark & \cmark & \cmark & \cmark \\
MRAG-Bench~\cite{humrag} & \xmark & \xmark & 1 & \xmark & \cmark & \xmark & \cmark & \xmark \\
ViQuAE~\cite{lerner2022viquae}     & \xmark & \xmark & 1 & \xmark & \cmark & \xmark & \xmark & \xmark \\
SK-VQA~\cite{su2024sk}     & \xmark & \xmark & 1 & \xmark & \cmark & \xmark & \cmark & \xmark \\
MMDocRAG~\cite{dong2025benchmarking}   & \xmark & \xmark & 1 & \xmark & \xmark & \cmark & \cmark & \cmark \\
\rowcolor{ourlightgray}
\ours      & \cmark & \cmark & \textbf{4} & \cmark & \cmark & \cmark & \cmark & \cmark \\
\bottomrule
\end{tabular}
\caption{Comparison of \ours~with current KB-VQA benchmarks. Our benchmark is the only one that explicitly evaluates hallucination and deflection across four retrieval scenarios: parametric (no additional context), oracle (gold evidence only), realistic (mixed positive and negative evidence), and adversarial (only negative contexts).}
\label{tab:benchmark_comparison}
\end{table*}

With this benchmark we can address a central question: \emph{can LVLMs deflect when they do not have enough information?} Our results show that the answer is still no. Experiments with 20 state-of-the-art LVLMs indicate that even the strongest proprietary systems fail to deflect reliably, often hallucinating when provided incomplete or distracting knowledge.

We note that \ours~is designed as a reusable and extensible framework rather than a one-off dataset. Future versions can rerun the curation pipeline to incorporate additional KB-VQA sources and include stronger models to filter parametrically-answerable questions. This ensures \ours~will continue to provide controlled conditions for measuring hallucination and deflection behavior under realistic retrieval noise, while preserving comparability through fixed, versioned releases. To our knowledge, it is the first KB-VQA benchmark to combine multimodal retrieval with explicit evaluation of hallucination and deflection, while also providing a dynamic foundation that can evolve alongside LVLMs. In summary, our contributions are threefold: 
\begin{itemize}
    \item We propose a dynamic curation pipeline that filters out parametrically solvable samples, ensuring that retained questions remain genuinely retrieval-dependent as models improve.
    \item We use this pipeline to build \ours, a collection of 2,775 samples covering diverse multimodal retrieval scenarios, each paired with both gold and distractor contexts. 
    \item We define a fine-grained evaluation protocol with four complementary scenarios: parametric, oracle, realistic, and adversarial that disentangle memorization from retrieval robustness and explicitly measure hallucination versus deflection.
\end{itemize}

\section{Related Work}

\paragraph{Knowledge-Based VQA Benchmarks.}
Research on Knowledge-based Visual Question Answering (KB-VQA) \cite{qiu2024snapntell, wu2025visual, shah2019kvqa, su2024sk} has progressed from early entity-centric datasets such as OK-VQA~\cite{marino2019ok}, A-OKVQA~\cite{schwenk2022okvqa}, and ViQuAE~\cite{lerner2022viquae}, which assumed every question was answerable and discouraged abstention. InfoSeek~\cite{chen2023can} and Encyclopedic-VQA~\cite{mensink2023encyclopedic} introduced retrieval pipelines, while WebQA~\cite{chang2022webqa} emphasized multi-hop reasoning. However, these benchmarks remained primarily text-centric, where external knowledge was supplied only as passages, without visual retrieval or multimodal distractors. Evaluation thus reduced to text-grounding accuracy, treating abstention as incorrect. More recent multimodal retrieval benchmarks~\cite{wu2025visual, dong2025benchmarking} such as MRAG-Bench~\cite{humrag} incorporate visual contexts, moving toward realistic multimodal information needs. However, they still lack explicit mechanisms for unanswerable queries or for distinguishing hallucination from deflection.

\paragraph{Hallucination and Deflection Benchmarks.}
Parallel efforts have begun probing reliability beyond accuracy. Vision-only datasets such as HaloQuest~\cite{wang2024haloquest} and AMBER~\cite{wang2023amber} diagnose hallucinations from mis-grounded visual evidence, while text-only sources~\cite{yang2024crag, wei2024measuring} such as MultiHop-RAG~\cite{tangmultihop} and GaRaGe~\cite{sorodoc2025garage} include unanswerable queries and explicitly reward abstention. However, they remain unimodal and cannot capture conflicts between retrieved text and images.

\paragraph{Retrieval-Augmented Multimodal Models.}
Recent advances in KB-VQA have focused on developing specialized architectures~\cite{deng2025muka, yuan2025mkg, xuan2024lemma, chen2022murag} -- including hierarchical retrieval~\cite{caffagni2024wiki}, re-ranking~\cite{yan2024echosight}, and adaptive control mechanisms~\cite{cocchi2025augmenting, asai2024self}. These works highlight the centrality of retrieval for LVLMs but generally assume high retrieval quality and measure only accuracy, leaving reliability under noisy or adversarial contexts underexplored.

Table~\ref{tab:benchmark_comparison} compares \ours~with other representative KB-VQA benchmarks. Existing resources focus narrowly on accuracy, omit distractors, or rely on uncontrolled online retrieval, which, while realistic, makes evaluation difficult to control and reproduce over time. In contrast, \ours~explicitly evaluates hallucination and deflection across four complementary scenarios, with pre-retrieved gold and distractor contexts that preserve benchmark stability as LVLMs evolve. To our knowledge, it is the first KB-VQA framework to address hallucination and deflection in multimodal retrieval.

\begin{figure*}[!ht]  \includegraphics[width=\linewidth]{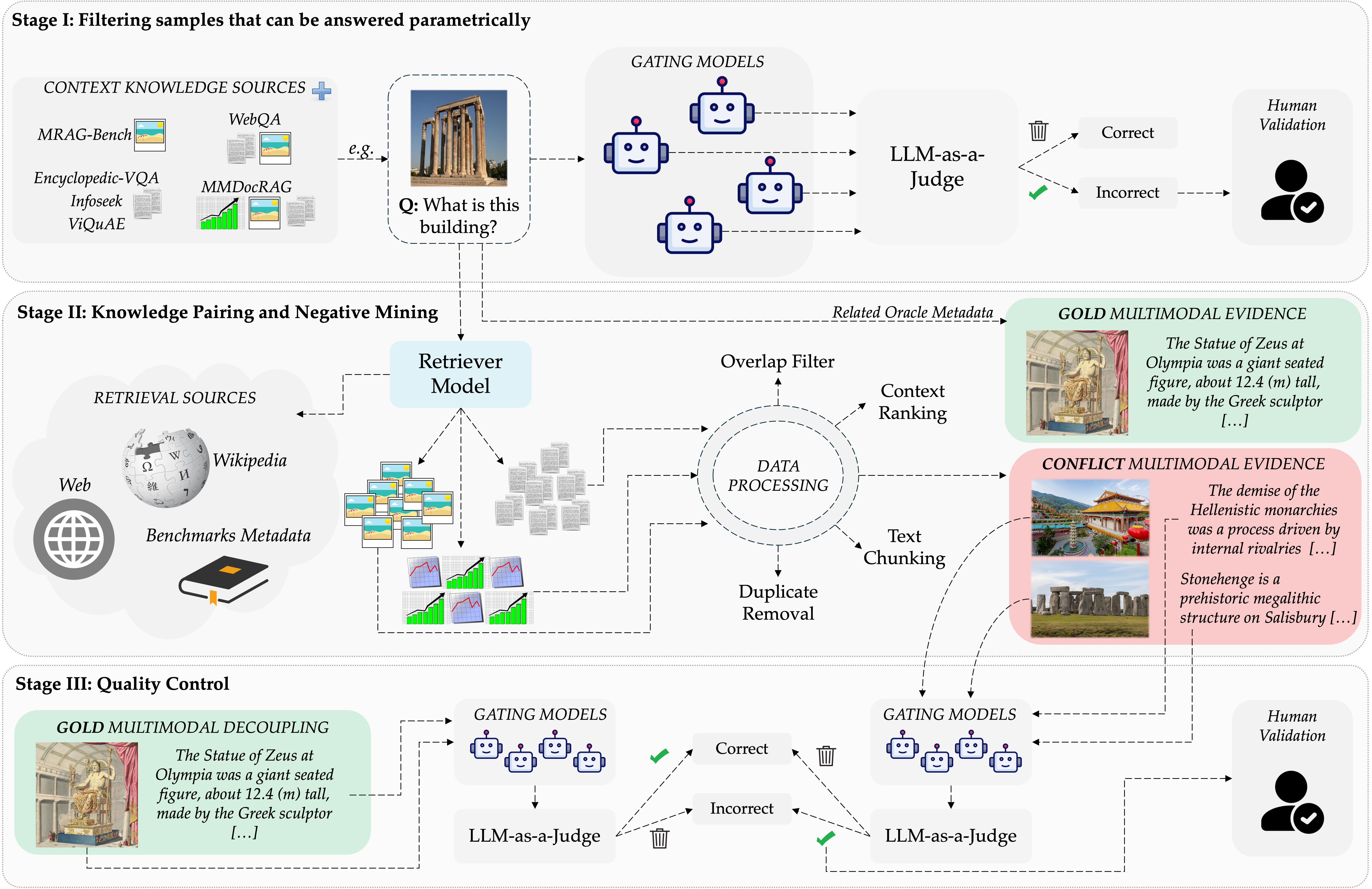}
  \caption{Pipeline of \ours. Starting from 6 benchmarks, we apply parametric filtering to remove query-solvable samples (\textsc{stage I}), retrieve negative (multimodal) contexts via different indices (\textsc{stage II}), then perform oracle filtering to eliminate false positive contexts (where gating models fail) and unreliable negative contexts (where gating models succeed) (\textsc{stage III}).}
  \label{fig:pipeline}
\end{figure*}

\section{Multi-Modal Deflection Benchmark}
\label{sec:method}

Unlike static datasets, \ours~is built through a dynamic pipeline parameterized by strong open-weight models and an external evaluator, ensuring adaptability as model capabilities evolve. As shown in Figure~\ref{fig:pipeline}, the pipeline consists of three stages: (i) filtering parametrically-answerable samples, (ii) augmenting retained samples with gold and distractor knowledge contexts, and (iii) quality control to guarantee both solvability and difficulty.

\subsection{Stage I: Filtering samples that can be answered parametrically}
Let $\mathcal{S}_0=\{x\}$ denote the pool of raw samples collected from heterogeneous KB-VQA sources. A KB-VQA instance is $x=(q,v,\mathcal{K},a)$, where $q$ is a question, $v$ is an optional image, $\mathcal{K}$ is candidate knowledge, and $a$ is the gold answer. We denote by $\mathcal{G}=\{G_1,\dots,G_n\}$ a set of \textit{gating models}\footnote{We set $\mathcal{G} = \{$\textsc{Gemma3-27B}, \textsc{Qwen-2.5-VL-32B}, \textsc{InternVL3-38B}, \textsc{VL-Rethinker-72B}\}}.
Not all sources provide visual inputs, so $v$ may be empty. This reflects the diversity of existing KB-VQA datasets, which range from purely text-based queries to fully multimodal questions requiring both text and images.

\begin{table}[t]
\centering
\footnotesize
\resizebox{0.48\textwidth}{!}{%
\begin{tabular}{lcccc}
\toprule
\textbf{Source} & \textbf{Samples} & \textbf{Filtered (\%)} & \textbf{Query} & \textbf{Context} \\
\midrule
InfoSeek          & 1646  & 5.6\% & Multi & Textual \\
WebQA             & 826  & 16.6\% & Textual & Multi \\
E-VQA             & 185  & 4.9\% & Multi & Textual \\
MMDocRAG          & 58  & 1.5\% & Textual & Multi \\
MRAG-Bench        & 45  & 3.3\% & Multi & Visual \\
ViQuAE            & 15  & 1.2\% & Multi & Textual \\
\midrule
\textbf{Total}    & 2,775 & 5.1\% & Multi & Multi \\
\bottomrule
\end{tabular}
}
\caption{Per-source breakdown of our dataset. We report the number of retained samples, filtering rate, and dominant modality for each source. The ``Filtered'' column indicates the percentage of samples retained from the original source after applying our filtering criteria.}

\label{tab:source_breakdown}
\end{table}

\begin{table}[t]
\centering
\small
\begin{tabular}{lc}
\toprule
\textbf{Statistic} & \textbf{Value} \\
\midrule
Total samples & 2,775 \\
Unique questions & 1,246 \\
Unique query images & 2,717 \\
Avg. question length (words) & 12.5 \\
Avg. gold contexts per sample & 1.9 \\
Avg. negative contexts per sample & 11.8 \\
\bottomrule
\end{tabular}
\caption{Global statistics of \ours. The dataset combines gold and negative contexts across text and vision, ensuring multimodal coverage and non-triviality.}
\label{tab:global_stats}
\end{table}

To eliminate parametrically solvable instances, each $G_j$ is queried in the \textbf{parametric setting with visual query} (hereinafter, the parametric setting), where the model receives the question $q$ together with the associated image $v$ when available, but no external knowledge: $\hat{a}_j = G_j(q,v).$
Responses are judged by an external evaluator $E$: we use GPT-4o as a judge with the \textsc{SimpleQA} prompt~\cite{wei2024measuring}, which assigns one of three labels: $E(q,\hat{a}_j,a) \in \{\texttt{CORRECT},\,\texttt{INCORRECT},\,\texttt{NOT ATTEMPTED}\}$\footnote{We map these labels to our evaluation metrics as follows: \texttt{CORRECT} $\rightarrow$ Accuracy, \texttt{INCORRECT} $\rightarrow$ Hallucination, \texttt{NOT ATTEMPTED} $\rightarrow$ Deflection.}.

We retain only samples unsolvable by all gating models:
\begin{equation}
\begin{gathered}
\mathcal{R}(x)=1 \;\;\Longleftrightarrow \\
\forall G_j \in \mathcal{G}, \; E(q, G_j(q,v),a)=\texttt{INCORRECT.}
\end{gathered}
\end{equation}

This produces $\mathcal{S}_1 \subseteq \mathcal{S}_0$, a set requiring external knowledge to solve.

\subsection{Stage II: Knowledge Pairing and Negative Mining}
Each retained sample $x \in \mathcal{S}_1$ is paired with gold knowledge $K(x)^+$ and distractors $K(x)^-$. If negatives exist in the source dataset, we adopt them; otherwise, we mine additional distractors via retrieval. For textual negatives, we construct a Wikipedia index $\mathcal{W_T}$ and retrieve candidate pages using EVA-CLIP (denoted as $R^{\phi}$, where $\phi$ represents the similarity function). These pages are then segmented into passages and re-ranked using Contriever~\cite{izacardunsupervised}; for $x=(q,v,\mathcal{K}, a)$:
\begin{equation}
\begin{gathered}
C(x) = \text{Chunk}(\text{top-}10(R^{\phi}(v; \mathcal{W_T}))) \\
K(x)^{-\text{txt}} = \text{Rerank}(C(x)) \setminus K(x)^+
\end{gathered}
\end{equation}

For visual negatives, we build an index $\mathcal{W_I}$ of query and metadata images, retrieving the top-10 most similar non-gold images:
\begin{equation}
K(x)^{-\text{img}} = \text{top-}10(R^{\phi}(v; \mathcal{W_I})) \setminus K(x)^+.
\end{equation}

\subsection{Stage III: Quality Control}
We apply two checks. First, solvability: if all gating models fail even with gold evidence, the sample is discarded:
\begin{equation}
\forall G_j \in \mathcal{G}, \quad E(q, G_j(q,v,K^+), a) = \texttt{INC}.
\end{equation}
Second, distractor validity: if any gating model exploits a distractor $k^-$ to answer correctly, that distractor is removed:
\begin{equation}
\exists G_j \;\; \text{s.t.} \;\; E(q, G_j(q,v,k^-),a) = \texttt{CORRECT}.
\end{equation}
Finally, we enforce a minimum of $K_{\min}=5$ negatives per sample, yielding the final benchmark:
\begin{equation}
\mathcal{S}^* = \{s,(K(x)^+,K(x)^-)\},
\end{equation}
and we randomly shuffle the order of positive and negative contexts before presenting them to models during evaluation, to simulate realistic retrieval scenarios.

\subsection{Benchmark Definition}
The final benchmark comprises 2,775 curated samples, each paired with gold and distractor passages. All contexts are pre-retrieved, ensuring reproducibility without external databases while reflecting real-world retrieval noise.

Table~\ref{tab:source_breakdown} summarizes \ours, reporting per-source retention, filtering ratio, dominant modality, and average of multimodal contexts. Complementary statistics are shown in Table~\ref{tab:global_stats}, including unique questions and images, average question length, and prevalence of multimodal distractors. Full distributions are provided in Appendix~\ref{sec:appendix_distribution}.

\subsection{Evaluation Scenarios}
We define four scenarios by controlling the context $\mathcal{Z}(x)$:  
\textbf{Parametric} ($\mathcal{Z}_P(x)=\emptyset$) forces reliance on parametric knowledge only, while still grounding on the input image $v$ when available.
\textbf{Oracle} ($\mathcal{Z}_O(x)=K^+$) supplies only gold knowledge.  
\textbf{Realistic} ($\mathcal{Z}_R(x)=K^+ \cup K^-$) mixes gold and distractors.  
\textbf{Adversarial} ($\mathcal{Z}_A(x)=K^-$) includes only distractors. Ideal behavior varies by scenario: \textbf{Parametric} should yield minimal accuracy, confirming our controlled strict-RAG setup; \textbf{Oracle} should achieve high accuracy given perfect evidence; \textbf{Realistic} should maintain high accuracy despite distractors; \textbf{Adversarial} should maximize deflection. Across all conditions, robust models should exhibit low hallucination rates. We acknowledge that some parametric accuracy is expected due to models' inherent knowledge.

Together, these scenarios span the full spectrum from knowledge-free to perfectly grounded to noisy and misleading contexts. Crucially, our framework implements a rigorous strict retrieval-augmented generation evaluation: we assess systems based on their factual accuracy and their response when confronted with inadequate or contradictory retrieved context. A trustworthy RAG system should recognize when to withhold answers due to poor evidence quality, rather than generate confident but unsubstantiated claims.

\begin{table*}[!ht]
\centering
\resizebox{\textwidth}{!}{%
\begin{tabular}{lcccccccccccc}
\toprule
\multirow{2}{*}{\textbf{Model}} &
\multicolumn{3}{c}{\textbf{Parametric}} &
\multicolumn{3}{c}{\textbf{Oracle}} &
\multicolumn{3}{c}{\textbf{Realistic}} &
\multicolumn{3}{c}{\textbf{Adversarial}} \\
\cmidrule(lr){2-4} \cmidrule(lr){5-7} \cmidrule(lr){8-10} \cmidrule(lr){11-13}
& Acc ↑ & Defl ↑ & Hall ↓ & Acc ↑ & Defl ↓ & Hall ↓ & Acc & Defl ↓ & Hall ↓ & Acc ↑ & Defl ↑ & Hall ↓ \\
\midrule
\rowcolor{ourlightgray}
\multicolumn{13}{c}{\textbf{Open Source Models}} \\
\midrule
LLaVA-OneVision & 4.7 & 4.0 & 91.3 & 54.9 & \textbf{3.5} & 41.6 & 34.0 & 10.8 & 55.2 & \textbf{4.5} & 14.5 & 81.0 \\
MiniCPM-V-4.5-8B & 4.6 & 3.2 & 92.2 & 65.3 & 8.9 & 25.8 & 44.3 & 16.3 & 39.4 & 1.9 & 46.2 & 51.9 \\
Keye-1.5-8B & 3.0 & 23.9 & 73.1 & 63.4 & 9.3 & 27.3 & 47.6 & 11.9 & 40.5 & 2.8 & 43.4 & 53.8 \\
InternVL3-9B & 3.1 & 9.6 & 87.3 & 61.2 & 7.7 & 31.1 & 45.1 & 10.3 & 44.6 & 2.3 & 35.9 & 61.8 \\
GLM-4.1V-9B-Thinking & 4.9 & 7.7 & 87.4 & 55.2 & 31.5 & 13.3 & 37.4 & 40.2 & 22.4 & 1.2 & 76.3 & 22.5 \\
Pixtral-12B & 4.6 & 19.4 & 76.0 & 62.7 & 6.8 & 30.5 & 42.6 & 12.6 & 44.8 & 2.3 & 40.9 & 56.8 \\
Ovis2-16B & 4.0 & 25.2 & 70.8 & 64.4 & 4.2 & 31.4 & \textbf{61.1} & \textbf{3.7} & 35.2 & 3.9 & 24.2 & 71.9 \\
Mistral-Small-3.1-24B-Instruct & 2.5 & \textbf{70.9} & \textbf{26.6} & 42.6 & 47.1 & \textbf{10.3} & 23.5 & 61.6 & \textbf{14.9} & 0.6 & \textbf{83.8} & \textbf{15.6} \\
Aria-25B & 4.4 & 31.0 & 64.6 & 62.3 & 10.2 & 27.5 & 39.5 & 21.4 & 39.1 & 2.9 & 40.7 & 56.4 \\
\text{Gemma3-27B}$^{\star}$ & 4.6 & 7.1 & 88.3 & 59.5 & 16.1 & 24.4 & 42.9 & 22.6 & 34.5 & 1.5 & 59.1 & 39.4 \\
\text{Qwen2.5-VL-32B-Instruct}$^{\star}$ & 3.2 & 4.5 & 92.3 & 61.0 & 5.1 & 33.9 & 45.2 & 5.3 & 49.5 & 2.4 & 13.7 & 83.9 \\
Ovis2-34B & 4.5 & 21.7 & 73.8 & \textbf{66.5} & 6.7 & 27.8 & 49.1 & 7.6 & 43.3 & 3.2 & 38.7 & 58.1 \\
\text{InternVL3-38B}$^{\star}$ & 4.8 & 1.4 & 93.8 & 64.2 & 13.2 & 22.6 & 50.3 & 15.8 & 33.9 & 2.5 & 55.2 & 42.3 \\
\text{VL-Rethinker-72B}$^{\star}$ & \textbf{5.0} & 1.9 & 93.1 & 62.6 & 12.9 & 24.5 & 47.2 & 11.9 & 40.9 & 2.3 & 42.3 & 55.4 \\
\midrule
\rowcolor{ourlightgray}
\multicolumn{13}{c}{\textbf{Closed Source Models}} \\
\midrule
Claude-Sonnet-4 & 9.7 & 31.8 & 58.5 & 47.5 & 42.6 & 9.9 & 33.3 & 56.2 & 10.5 & 0.9 & 87.0 & 12.1 \\
Claude-Opus-4 & 6.7 & \textbf{65.0} & \textbf{28.3} & 49.1 & 41.7 & \textbf{9.2} & 32.1 & 59.4 & \textbf{8.5} & 0.7 & \textbf{88.3} & \textbf{11.1} \\
Gemini-2.5-Flash & 17.5 & 11.7 & 70.8 & 58.8 & 27.9 & 13.3 & 47.0 & 32.3 & 20.7 & 1.4 & 74.5 & 24.1 \\
Gemini-2.5-Pro & \textbf{27.0} & 7.5 & 65.5 & 59.8 & 26.3 & 13.9 & 51.0 & 28.5 & 20.5 & 1.7 & 76.1 & 22.2 \\
GPT-5 & 23.7 & 1.5 & 74.8 & \textbf{73.1} & \textbf{14.3} & 12.6 & \textbf{59.5} & \textbf{15.0} & 25.5 & \textbf{4.1} & 61.2 & 34.7 \\
\bottomrule
\end{tabular}
}
\caption{Performance of LVLMs across evaluation scenarios. We report Accuracy (Acc), Deflection Rate (Defl), and Hallucination Rate (Hall). The number of negatives is fixed at 2 to match the average number of gold contexts per sample, ensuring a balanced ratio of positive and negative evidence. $^{\star}$ marks selected gating models.}
\label{tab:main_results}
\end{table*}

\section{Benchmark Evaluation}
We evaluate across all scenarios using GPT-4o as a judge with the \textsc{SimpleQA} protocol, following prior works~\cite{wei2024measuring, haas2025simpleqa}. To verify robustness, we also conducted a human validation study, which showed $>92\%$ agreement and Cohen’s $\kappa{=}0.91$ across models of different scales (Appendix~\ref{sec:appendix_human_validation}). 

Since models are explicitly instructed to rely solely on provided evidence (\textit{i.e.}, a strict RAG setting), we assume all incorrect responses as hallucinations. We acknowledge that in the Realistic scenarios some \textsc{Incorrect} predictions may correspond to misgrounding (\textit{e.g.}, selecting a distractor) rather than unsupported fabrication. Nevertheless, we follow prior practice in grouping them under hallucination for consistency and comparability across scenarios. This unified formulation provides a reproducible framework for knowledge-based VQA.

\subsection{Experimental Setup}
\label{sec:setup}
We design our experiments to address four research questions. First, we ask whether current LVLMs can deflect reliably under different knowledge conditions (Sec.~\ref{sec:main_exp}). 
Second, we analyze how modality interacts with reliability (Sec.~\ref{sec:ablation_neg}).

Third, we study how prompting strictness influences the deflection--hallucination trade-off (Sec.~\ref{sec:ablation_severity}). Finally, we evaluate robustness to retrieval noise (Sec.~\ref{sec:ablation_retrieval}).

We evaluate 20 representative LVLMs, spanning both open-weight and proprietary systems. 
On the open-weight side, we include multiple variants from major families: \textsc{LLaVA-OneVision}~\cite{lillava}, \textsc{MiniCPM-4.5-8B}~\cite{yao2024minicpm}, \textsc{Keye-1.5}~\cite{yang2025kwai}, \textsc{InternVL3} (9B, 38B)~\cite{zhu2025internvl3}, \textsc{GLM-4.1V-9B}~\cite{hong2025glm}, \textsc{Pixtral}~\cite{agrawal2024pixtral}, \textsc{Ovis2} (16B, 34B)~\cite{lu2024ovis}, \textsc{Mistral-Small-3.1-24B-Instruct}, \textsc{Aria}~\cite{li2024aria}, \textsc{Gemma3-27B}~\cite{team2025gemma}, \textsc{Qwen-2.5-VL-32B}~\cite{Qwen2.5-VL} and \textsc{VL-Rethinker}~\cite{wang2025vl}, ranging from 7B to 72B parameters. 

For proprietary systems, we test \textsc{Claude-4 (Sonnet, Opus)}, \textsc{GPT-5} and \textsc{Gemini-2.5 (Flash, Pro)}~\cite{comanici2025gemini}, representing the current frontier of commercial LVLMs.  
All open-weight models are run with the vLLM framework\footnote{\url{https://docs.vllm.ai}}, using top-$p=1.0$, repetition penalty $1.05$, and temperature $0.2$. 
Closed-source APIs are queried with their default decoding settings. Unless otherwise specified, all experiments use the moderate prompting strictness (see Appendix~\ref{sec:appendix_prompts} for full prompt details).

\subsection{Deflection under Knowledge Conditions}
\label{sec:main_exp}
Table~\ref{tab:main_results} compares open- and closed-weight models across the four scenarios.  
In the \textbf{parametric} setting, where only the question and image are available, accuracy is near-zero for open-weight systems, confirming that \ours\ successfully filters out parametrically answerable questions. The majority of models attempt to answer most questions rather than abstain, with deflection rates below $35\%$. The only exception is \textsc{Mistral-Small-3.1}, which deflects on $70.9\%$ of queries. Proprietary systems behave differently: \textsc{Gemini-2.5-Pro} and \textsc{GPT-5} obtain accuracies of $27.0\%$ $23.7\%$, respectively, likely reflecting training-set contamination.

Providing \textbf{oracle} knowledge boosts accuracy substantially, with several open-weight models surpassing $60\%$ (\textsc{Ovis2-34B} $66.5\%$), and even outperforming \textsc{Claude-Opus-4} ($49.1\%$). Hallucination, however, remains strikingly high: \textsc{LLaVA-OneVision} hallucinates on $41.6\%$ of queries despite being given the correct evidence, showing that grounding, not retrieval, is the primary bottleneck.

In the \textbf{realistic} setting, where gold evidence is mixed with distractors, accuracy drops by $10$–$20$ points and hallucination often exceeds $40\%$. For instance, \textsc{Pixtral-12B} falls from $62.7\%$ to $42.6\%$ accuracy, while even \textsc{GPT-5} hallucinates on $25.5\%$ of queries.  

Finally, in the \textbf{adversarial} setting with only distractors where the desired behavior is near-total abstention, \textsc{Claude-Opus-4} achieves $88.3\%$ deflection and \textsc{Mistral-Small-3.1} reaches $83.8\%$. Some accuracy remains possible as models may still possess parametric knowledge to answer correctly despite misleading context.
Notably, consistent with the parametric setting, most open-weight systems attempt to answer questions rather than deflect, with \textsc{Qwen-2.5-VL} providing incorrect responses on $83.9\%$ of queries.

Overall, no model achieves balanced reliability: \textsc{Mistral} over-deflects, \textsc{Ovis2} excels only with clean evidence, and \textsc{Claude} sacrifices oracle accuracy for adversarial robustness. These contrasts illustrate the unique value of \ours\ in exposing hallucination–deflection trade-offs that remain invisible when measuring accuracy alone.

\begin{table}[!ht]
\centering
\resizebox{\linewidth}{!}{%
\begin{tabular}{lccccc}
\toprule
\multirow{2}{*}{\textbf{Model}} & \multicolumn{2}{c}{\textbf{Context}} & \multirow{2}{*}{Acc ↑} & \multirow{2}{*}{Defl ↓} & \multirow{2}{*}{Hall ↓} \\
\cmidrule(lr){2-3}
 & Positive & Negative &  &  &  \\
\midrule
\multirowcell{6}{Keye-1.5-8B} 
& \multirowcell{3}{Textual} 
    & MM   & 80.1 & 3.3 & 16.6 \\
&   & Visual  & 82.1 & 2.7 & 15.2 \\
&   & Textual & 82.5 & 3.3 & 14.2 \\
\cmidrule(lr){2-6}
& \multirowcell{3}{Visual}  
    & MM   & 1.0  & 82.5 & 16.5 \\
&   & Visual  & 9.4  & 19.2 & 71.4 \\
&   & Textual & 0.9  & 82.8 & 16.3 \\
\midrule
\multirowcell{6}{Mistral-Small-3.1} 
& \multirowcell{3}{Textual} 
    & MM   & 60.2 & 31.5 & 8.3 \\
&   & Visual  & 69.4 & 23.6 & 7.0 \\
&   & Textual & 53.8 & 39.4 & 6.8 \\
\cmidrule(lr){2-6}
& \multirowcell{3}{Visual}  
    & MM   & 0.8  & 98.1 & 1.1 \\
&   & Visual  & 3.8  & 93.2 & 3.0 \\
&   & Textual & 1.1  & 97.4 & 1.5 \\
\midrule
\multirowcell{6}{Claude-Opus-4} 
& \multirowcell{3}{Textual} 
    & MM   & 82.8 & 9.8 & 7.4 \\
&   & Visual  & 84.3 & 9.6 & 6.0 \\
&   & Textual & 80.8 & 13.0 & 6.2 \\
\cmidrule(lr){2-6}
& \multirowcell{3}{Visual}  
    & MM   & 18.8 & 72.9 & 8.3 \\
&   & Visual  & 35.3 & 50.8 & 13.9 \\
&   & Textual & 28.9 & 63.2 & 7.9 \\
\bottomrule
\end{tabular}
}
\caption{Effect of negative-context modality in the \textbf{Realistic} scenario. Results reveal a language-over-vision bias: models handle visual distractors well when positives are textual, but collapse when textual distractors interfere with visual positives, demonstrating that misleading text systematically overrides correct visual evidence.
}
\label{tab:ablation_multimodal}
\end{table}

\subsection{Impact of Negative Context}
\label{sec:ablation_neg}
In Table~\ref{tab:ablation_multimodal} we analyze model behavior in the \textbf{Realistic} scenario when the modality of negative contexts varies across textual, visual, or both. We evaluate 530 examples with textual positives and 266 examples with visual positives\footnote{Sample sizes differ because not all examples support both multimodal positive and negative contexts due to source limitations and question formulation constraints.}.
Notably, we observe a consistent drop in overall performance when moving from textual to visual positive contexts across all models, indicating that LVLMs struggle more with visual knowledge grounding than textual knowledge.
When positives are textual, models generally handle visual distractors well: \textsc{Claude-Opus-4} maintains $84.3\%$ accuracy with visual noise, while \textsc{Keye-1.5-8B} even improves slightly ($82.1\%$ vs.\ $80.1\%$ with mixed noise). 
In contrast, \textsc{Mistral-Small-3.1} becomes more conservative, dropping from $69.4\%$ accuracy with visual distractors to $60.2\%$ with mixed, while deflection rises from $23.6\%$ to $31.5\%$. The picture reverses when positives are visual: textual distractors dominate. Both \textsc{Keye-1.5-8B} and \textsc{Mistral-Small-3.1} collapse to near-zero accuracy ($\leq 1\%$) with deflection exceeding $80\%$, effectively refusing to answer once misleading text is present. Even \textsc{Claude-Opus-4} falls from $35.3\%$ accuracy with visual distractors to $28.9\%$ with textual ones. Multimodal distractors produce intermediate results but still degrade reliability. 

Crucially, this asymmetry reverses the expected grounding hierarchy: text dominates vision even when vision carries the gold evidence. The presence of misleading text systematically overrides correct visual signals, reducing accuracy across all architectures. Multimodal distractors yield intermediate degradation, reflecting partial—but still language-weighted—fusion. These results expose a persistent \textbf{language-over-vision bias} in current LVLMs: they rely on textual priors even under oracle visual grounding. This failure mode highlights the limits of current alignment training and underscores the need for \textit{retrieval-aware grounding} strategies that weight evidence by modality reliability rather than surface plausibility.

\subsection{Effect of Prompt Strictness and Trade-offs}
\label{sec:ablation_severity}

\begin{figure*}[!t]
\centering
  \includegraphics[width=\linewidth]{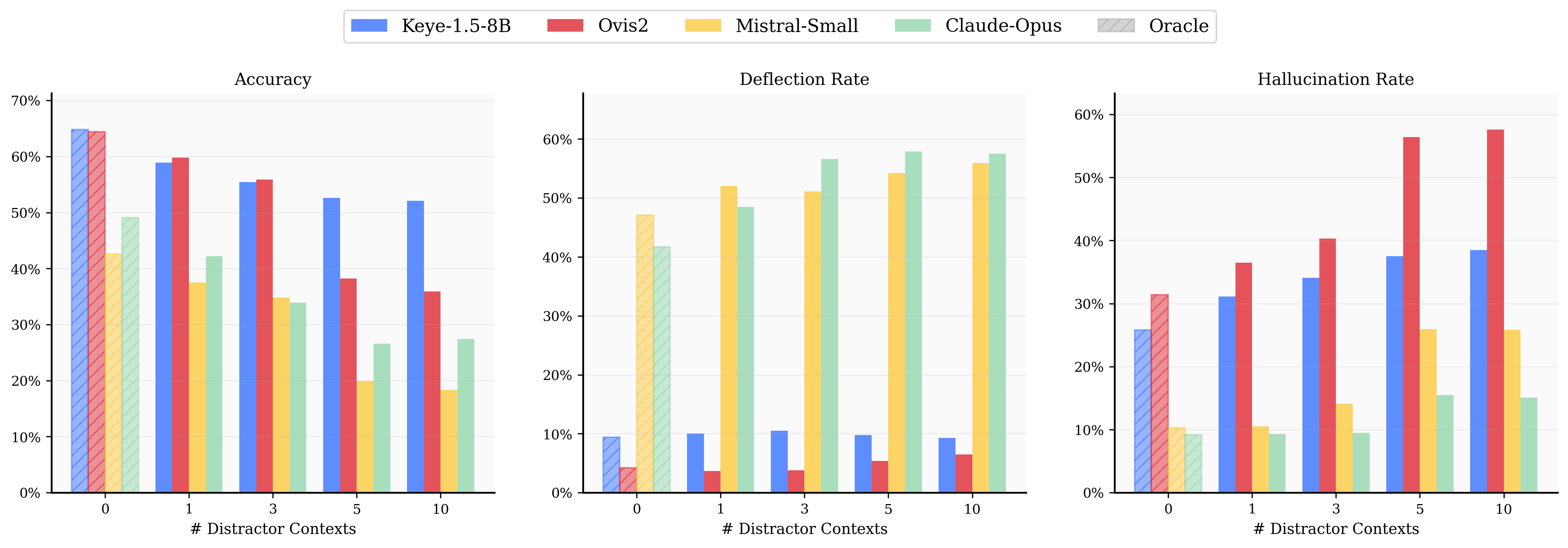}
  \caption{Effect of distractor quantity in the \textbf{Realistic} scenario, including \textbf{Oracle} results (0 distractors) for reference. Accuracy declines and hallucination rises as negatives increase, while deflection shows limited improvement.}
  \label{fig:ablation_context}
\end{figure*}

We test whether increasing the strictness of refusal instructions affects the balance between deflection and hallucination. 
Table~\ref{tab:table_severity} reports results for soft, moderate, and severe prompts (see Appendix~\ref{sec:appendix_prompts} for full prompt text) in the realistic and adversarial scenarios. 
These prompts differ only in how forcefully they direct the model to abstain when uncertain: 
the soft version politely suggests caution; the moderate version explicitly recommends abstention if unsure; 
and the strict version mandates refusal unless the answer is fully supported. Results show a clear but uneven effect. In the adversarial setting, where the target behavior is near-total abstention, stronger instructions substantially increase deflection. For instance, \textsc{Mistral-Small-3.1} jumps from $57.6\%$ to $98.2\%$ deflection with severe prompting, nearly eliminating hallucinations. Proprietary systems such as \textsc{Claude-Opus-4} also benefit, though less dramatically ($87.0\rightarrow 92.2\%$). 
By contrast, in the realistic setting where gold evidence is mixed with distractors, severity introduces a trade-off: hallucination decreases, but accuracy collapses. For example, \textsc{Keye-1.5-8B} drops from $58.7\%$ to $36.2\%$ accuracy, while \textsc{Mistral} falls from $57.7\%$ to $9.2\%$ accuracy under severe prompting, reflecting excessive caution. Overall, prompt engineering is a powerful but blunt tool for the prompts we tested. 
Strong refusal instructions enforce abstention in adversarial cases but cause over-deflection in realistic ones, undermining accuracy even when valid evidence is present. While model-specific prompt optimization might achieve better accuracy-deflection trade-offs, our results suggest that distinguishing genuine uncertainty from noisy retrieval remains difficult and cannot be easily solved by prompting alone.

\subsection{Robustness to Retrieval Noise}
\label{sec:ablation_retrieval}
To test how LVLMs cope with noise, we vary the number of negatives from $1$ to $10$ in the \textbf{realistic} scenario (Figure~\ref{fig:ablation_context}). Results reveal a sharp vulnerability to distractor density. Accuracy steadily declines while hallucination rises, even for strong systems. For instance, \textsc{Ovis2-16B} drops from $59.8\%$ accuracy with one distractor to $35.9\%$ with ten, with hallucination nearly doubling ($36.5\% \rightarrow 57.6\%$). Proprietary models are more resilient but not immune: the accuracy of \textsc{Claude-Opus-4} still falls from $42.2\%$ to $27.4\%$, while its hallucination remains between $9$–$15\%$. Crucially, deflection does not increase enough to compensate. \textsc{Mistral-Small-3.1-24B}, for example, raises its deflection rate only marginally ($52.0 \rightarrow 55.9\%$) while accuracy collapses from $37.5$ to $18.3\%$). This pattern shows that models prefer hallucination over abstention when retrieval is cluttered. 

\begin{table}[!ht]
\centering
\resizebox{\columnwidth}{!}{%
\begin{tabular}{lccccccc}
\toprule
\multirow{2}{*}{\textbf{Model}} & 
\multirow{2}{*}{\textbf{Severity}} & 
\multicolumn{3}{c}{\textbf{Realistic}} &
\multicolumn{3}{c}{\textbf{Adversarial}} \\
\cmidrule(lr){3-5} \cmidrule(lr){6-8}
& & Acc ↑ & Defl ↓ & Hall ↓ & Acc ↑ & Defl ↑ & Hall ↓ \\
\midrule
\multirow{4}{*}{Keye-1.5-8B} 
& None     & 58.7 & 4.7 & 36.6 & 4.1 & 21.0 & 74.9 \\
& Soft     & 56.9 & 8.8 & 34.3 & 3.2 & 37.2 & 59.7 \\
& Moderate & 56.0 & 9.8 & 34.2 & 3.1 & 43.8 & 53.1 \\
& Severe   & 36.2 & 42.9 & 20.9 & 1.5 & 77.2 & 21.3 \\
\cmidrule{1-8}
\multirow{4}{*}{Ovis2-16B} 
& None        & 57.9 & 1.0 & 41.1 & 4.6 & 5.7 & 89.7 \\
& Soft     & 59.4 & 2.4 & 38.2 & 3.8 & 13.7 & 82.5 \\
& Moderate & 61.1 & 3.7 & 35.2 & 3.9 & 24.2 & 71.9 \\
& Severe   & 50.9 & 21.4 & 27.7 & 1.7 & 62.3 & 36.0 \\
\cmidrule{1-8}
\multirow{4}{*}{Mistral-Small-3.1} 
& None  & 57.7 & 16.9 & 25.4 & 2.2 & 57.6 & 40.2 \\
& Soft     & 51.0 & 27.3 & 21.7 & 1.7 & 69.3 & 29.0 \\
& Moderate & 23.5 & 61.6 & 14.9 & 0.6 & 83.8 & 15.6 \\
& Severe   & 9.2 & 89.9 & 0.9 & 0.2 & 98.2 & 1.6 \\
\cmidrule{1-8}
\multirow{4}{*}{Claude-Opus-4} 
& None & 37.7 & 52.0 & 10.3 & 0.6 & 87.5 & 11.9 \\
& Soft     & 38.1 & 52.0 & 9.9 & 0.8 & 87.0 & 12.2 \\
& Moderate & 32.1 & 59.4 & 8.5 & 0.7 & 88.3 & 11.0 \\
& Severe   & 32.0 & 62.3 & 5.7 & 0.7 & 92.2 & 7.1 \\
\bottomrule
\end{tabular}
}
\caption{Effect of prompt strictness across \textbf{Realistic} and \textbf{Adversarial} scenarios for four representative LVLMs. Higher strictness increases deflection rates and reduces hallucination while decreasing accuracy in realistic scenarios.}
\label{tab:table_severity}
\end{table}

Overall, this experiment demonstrates that retrieval noise, not just retrieval quality, is a fundamental bottleneck. As soon as distractor density increases, both open-weight and proprietary models struggle to separate signal from noise, defaulting to hallucinations instead of deflection, even when explicitly prompted to deflect properly.

\section{Conclusion}
We introduced \ours, a benchmark for measuring how LVLMs balance accuracy, deflection, and hallucination under varying knowledge conditions. Results show that even top-performing systems hallucinate when evidence is incomplete and over-deflect when refusal cues are too strong, revealing a lack of calibrated confidence. Beyond static datasets, \ours~provides a flexible and model-agnostic framework that can evolve with stronger judging and gating models. By unifying KB-VQA and retrieval-augmented evaluation under a common protocol, it offers a step toward trustworthy multimodal reasoning—where models learn not only to answer correctly but also to know when not to answer.

\section{Limitations}
We acknowledge several limitations of our work. First, our evaluation relies on GPT-4o with the validated \textsc{SimpleQA} protocol as an automatic judge. This enables scalability and consistency, but it cannot fully replace human annotations, especially for nuanced multimodal reasoning. Second, the benchmark inherits biases from its source datasets, which are predominantly text-centric and may underrepresent knowledge that is primarily visual or domain-specific. 

Third, our experiments are limited to license-friendly or API-accessible models, which excludes certain families such as some LLaMA-based derivatives. Fourth, LVLMs may show stochastic variability across runs; due to inference costs we report results on a single pass per query, while acknowledging that repeated sampling could yield small fluctuations. Fifth, \ours\ focuses on short-form question answering; extending it to long-context reasoning or interactive dialogue remains open for future work. 
These limitations highlight opportunities for further research without diminishing our central finding that hallucination and deflection remain persistent challenges, especially for Retrieval Augmented Generation models.

Finally, due to the fact that the benchmark is constructed from pre-existing publicly available datasets, we cannot guarantee that offensive and harmful language and images are not included from the source datasets.

\section*{Acknowledgments}

Bill Byrne holds concurrent appointments as an Amazon Scholar and as Professor of Information Engineering at the University of Cambridge. This paper describes work performed at Amazon.

\bibliography{custom}

\clearpage

\appendix

\section*{Appendix}
\addcontentsline{toc}{section}{Appendix}

This appendix provides supplementary materials and extended analyses to complement the main paper. We first present a detailed comparison between our benchmark and prior multimodal question-answering datasets. We then discuss guidelines for refreshing the benchmark, analyze data composition and pipeline convergence, and compare GPT-4o–based automatic judgments with human annotations. Further sections include model-wise performance breakdowns, sensitivity to prompt strictness, implementation details, full prompt templates, and qualitative examples illustrating common failure modes. These materials are intended to ensure transparency, reproducibility, and a deeper understanding of our proposed benchmark.

\section{Extended Comparison with existing benchmarks}
Our benchmark draws on six publicly available datasets, each targeting complementary aspects of knowledge-intensive multimodal reasoning.

\textbf{InfoSeek}~\cite{chen2023can} is a large-scale knowledge-seeking VQA dataset built from Wikipedia. It combines a high-quality human-written set with over 1.3M automatically generated image–question–answer pairs, designed to test fine-grained factual queries about images that require external knowledge rather than surface-level recognition.

\textbf{Encyclopedic-VQA}~\cite{mensink2023encyclopedic} emphasizes fine-grained, encyclopedic knowledge. It contains over 220k unique question–answer pairs linked to 16.7k categories, each paired with multiple images. Questions are grounded in $\sim$2M Wikipedia articles, many requiring multi-hop reasoning and providing evidence attribution to specific passages.

\textbf{WebQA}~\cite{chang2022webqa} simulates real-world web search QA. It consists of $\sim$46k user-like queries paired with noisy pools of text snippets and images retrieved from the web, including hard distractors. Queries often require multi-hop reasoning across modalities, making it a challenging open-domain multimodal QA benchmark.

\textbf{MMDocRAG}~\cite{dong2025benchmarking} targets retrieval-augmented generation over long multimodal documents. It comprises $\sim$4k annotated QA pairs grounded in 222 documents across diverse domains, interleaving text with tables, charts, and figures. The benchmark evaluates whether models can retrieve fine-grained evidence and generate answers that may integrate both text and visual content.

\textbf{MRAG-Bench}~\cite{humrag} is a diagnostic benchmark for vision-centric retrieval-augmented QA. It includes 1.3k multiple-choice questions and a curated corpus of $\sim$9.6k images. Scenarios are designed around perspective changes and transformations, requiring retrieval of supporting images to answer questions that textual knowledge alone cannot resolve.

\textbf{ViQuAE}~\cite{lerner2022viquae} is a smaller-scale benchmark ($\sim$3.7k examples) focusing on entity-centric knowledge-based VQA. Questions originate from TriviaQA, with the entity mention replaced by an image. Each example links to the relevant Wikipedia page and section, requiring models to identify the entity visually and retrieve the corresponding fact.

These datasets span encyclopedic, web-based, document-level, vision-centric, and entity-centric settings, providing a diverse testbed for evaluating retrieval-augmented multimodal language models.

\section{Benchmark Evolution and Maintenance}
Although this release of \ours~is fixed and versioned for comparability, the underlying pipeline is designed for \textbf{controlled evolution}. Its modular structure supports the inclusion of new models, data sources, and evaluation criteria as the field advances. When stronger models reach high parametric accuracy, they can be incorporated into the gating pool to reapply filtering and remove samples that no longer require retrieval. This procedure keeps \ours~retrieval-dependent and behaviorally informative, while explicit versioning guarantees reproducibility across updates. Implementation details are provided in Section~\ref{sec:implementation}.

Beyond stability, a central strength of our framework lies in its \textbf{flexibility}. The pipeline is not tied to a fixed ensemble or annotation protocol but can be adapted to any retrieval-augmented benchmark. In practice, researchers can substitute the gating models with the strongest available candidates and use the most reliable automatic judge for evaluation. As new LVLMs and RAG systems emerge, this design remains future-proof, since it makes no assumptions about specific architectures or knowledge sources.

This flexibility also enables integration across diverse knowledge-grounded tasks. Any KB-VQA or RAG benchmark can be incorporated by re-running the same pipeline with updated models, ensuring consistent retrieval dependence and evaluation semantics. Moreover, because the framework explicitly enforces retrieval grounding, it harmonizes datasets that have historically differed in annotation style, distractor construction, or evaluation criteria.

We view \ours~as a \textbf{foundation for unified evaluation} in the KB-VQA community: a shared judge, a principled deflection-aware metric, and an ensemble of retrieval-dependent samples that evolve transparently with model progress. By decoupling benchmark content from model implementation, \ours~remains both reproducible and extensible, able to grow with the field while preserving comparability across generations of LVLMs.

\section{Data Distribution Analysis}
\label{sec:appendix_distribution}
\begin{figure*}[!ht]
  \includegraphics[width=\linewidth]{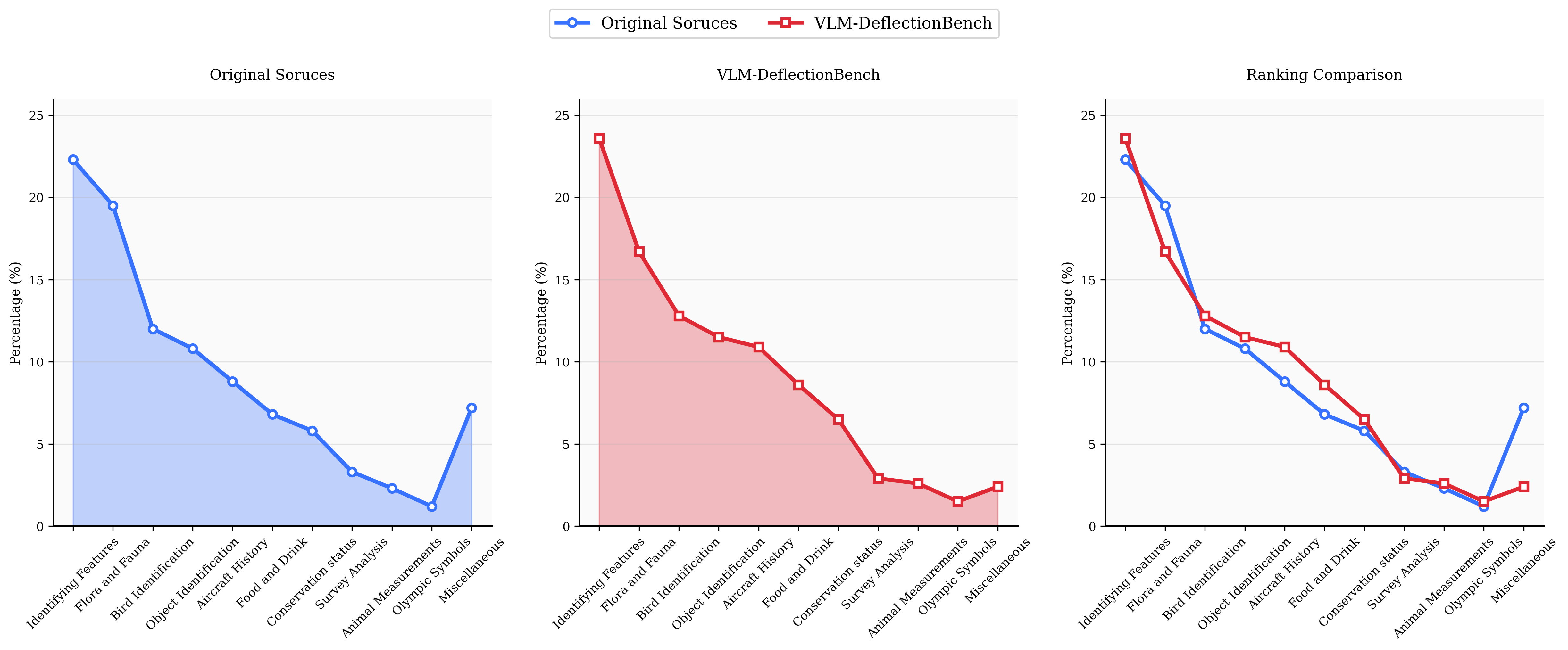}
  \caption{Comparison of category distributions before and after curation. (Left) Aggregate distribution of the six original source benchmarks. (Middle) Final distribution in \ours. (Right) Overlay showing relative shifts across categories.}
  \label{fig:appendix_distribution}
\end{figure*}

To assess whether our filtering pipeline alters the topical composition of the data, Figure~\ref{fig:appendix_distribution} compares sample distributions before and after curation, following~\citet{tang2025explorer}. We use \textsc{BERTopic} to cluster questions based on embeddings from \textsc{OpenAI text-embedding-3-large}, reduced via UMAP and grouped with HDBSCAN using cluster probability scores. For interpretability, we select ten representative questions per cluster and use GPT-4o to assign category names and brief descriptions (each defined by the query image and question). 
The left panel aggregates all six source benchmarks (\textsc{InfoSeek}, \textsc{Encyclopedic-VQA}, \textsc{WebQA}, \textsc{MMDocRAG}, \textsc{ViQuAE}, and \textsc{MRAG-Bench}); the middle panel shows the resulting distribution in \ours; and the right panel overlays the two. While relative weights shift slightly (\textit{e.g.}, \textit{Identifying Features} and \textit{Food and Drinks} become marginally more frequent), the overall long-tailed shape is preserved. In particular, core categories such as \textit{Object Identification} and \textit{Conservation Status} remain within one to two percentage points of their original prevalence. 
This analysis confirms that our filtering pipeline does not bias \ours~toward a few sources or topics but maintains a broad and diverse question distribution while enforcing stricter retrieval dependence and grounding fidelity.

\section{Pipeline Effectiveness and Convergence}
\label{sec:appendix_convergence}
\begin{figure*}[!ht]
  \includegraphics[width=\linewidth]{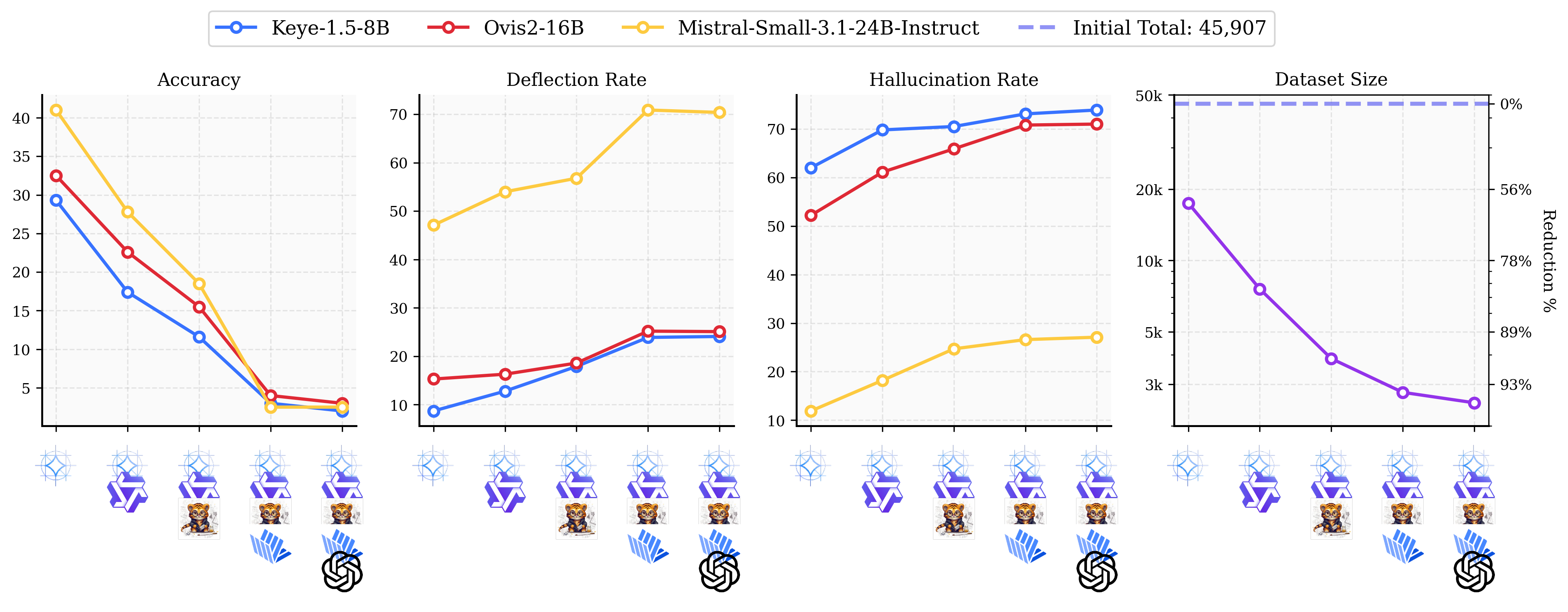}
  \caption{Performance on \textbf{Parametric Scenario} and dataset size as models are incrementally added to the filtering pipeline. Convergence is reached after four models, with over 90\% of samples removed.}
  \label{fig:appendix_convergence}
\end{figure*}

Figure~\ref{fig:appendix_convergence} illustrates how dataset composition evolves as successive filters are applied in our pipeline. To assess convergence, we monitor three representative LVLMs: \textsc{Keye-1.5-8B}, \textsc{Ovis2-16B}, and \textsc{Mistral-Small-3.1-24B-Instruct} under the parametric setting. As expected, accuracy decreases monotonically with each stage, indicating that questions solvable through parametric recall are progressively eliminated. At the same time, deflection increases, showing that the remaining samples require grounding in external evidence rather than model memory. Hallucination rates also rise, reflecting the removal of parametric instances and the retention of more challenging, knowledge-intensive ones.
The rightmost panel shows that over 90\% of the initial $\sim$45.9k pooled samples are filtered out, converging to approximately 2.7k. This reduction stems not only from eliminating parametrically answerable items but also from pruning residual noise—ambiguous queries, unanswerable cases even under gold evidence, and false negatives where distractors were mislabeled as positives. Convergence stabilizes after four models, as the inclusion of \textsc{GPT-4o} produces negligible additional filtering. We therefore define this point as the convergence threshold, balancing dataset compactness and source coverage. Despite this aggressive pruning, the final dataset preserves the relative distribution of sources from the original pool (see Figure~\ref{fig:appendix_distribution}), ensuring representativeness.
Overall, these trends confirm that the pipeline effectively suppresses parametric leakage, removes residual noise, and converges to a compact yet diverse dataset where success requires genuine retrieval-based reasoning rather than memorization. The complete workflow is formalized in Algorithm~\ref{appendix_algorithm}.

\begin{algorithm}[t]
\caption{Dynamic Curation Pipeline}
\KwIn{\var{Raw samples} $S_0$, \var{Gating models} $G$, \var{Textual KB} $K_{\text{txt}}$, \var{Visual KB} $K_{\text{img}}$, \var{Threshold} $K_{\min}$}
\KwOut{\var{Curated benchmark} $S^*$}

\BlankLine
$S_1 \gets \emptyset$ \;

\For{$s \in S_0$}{
    $decisions \gets \{ J(g(I_s, q_s)) : g \in G \}$ \;
    \If{$\text{all}(decisions = \const{INCORRECT})$}{
        $S_1 \gets S_1 \cup \{s\}$ \;
    }
}

\BlankLine
\For{$s \in S_1$}{
    $K_s^+ \gets$ \kw{gold contexts} \;
    $K_s^- \gets$ \kw{retrieve negatives from} $K_{\text{txt}}, K_{\text{img}}$ \;

    \If{$\nexists g \in G : J(g(I_s,q_s,K_s^+)) = \const{CORRECT}$}{
        \kw{discard} $s$; \textbf{continue}\;
    }

    \For{$n \in K_s^-$}{
        \If{$\exists g \in G : J(g(I_s,q_s,n)) = \const{CORRECT}$}{
            \kw{remove} $n$ \;
        }
    }

    \If{$|K_s^-| < K_{\min}$}{
        \kw{discard} $s$ \;
    }
    \Else{
        \kw{retain} $s$ with $(K_s^+, K_s^-)$ \;
    }
}
\Return $S^*$ \;
\label{appendix_algorithm}
\end{algorithm}

\section{Human Validation vs. GPT-4o Judgments}
\label{sec:appendix_human_validation}

To evaluate the reliability of GPT-4o as an automatic judge, we conducted a human validation study. We sampled 100 responses per model, balanced across the four evaluation scenarios, from four representative LVLMs spanning different scales: \textsc{Keye-1.5-8B} (small, open-weight), \textsc{Ovis2-16B} (mid-size, open-weight), \textsc{Mistral-3.1-24B-Instruct} (large, open-weight with refusal-oriented prompting), and \textsc{Claude-Opus-4} (frontier-level proprietary). In total, 400 responses were annotated by a human rater using the same \textsc{SimpleQA} rubric with three labels: \textsc{Correct}, \textsc{Incorrect}, and \textsc{Not\_Attempted}. 

We compared human and GPT-4o labels using two complementary metrics: (i) raw agreement (\%), the proportion of identical labels, and (ii) Cohen’s~$\kappa$, which accounts for chance agreement given the label distribution ($\kappa{=}1$ indicates perfect, $\kappa{=}0$ chance-level agreement). Results in Table~\ref{tab:human_validation} show consistently high alignment across models, with an average agreement of $92.6\%$ and $\kappa{=}0.91$, corresponding to \textit{almost perfect} agreement. 
Discrepancies were rare and largely involved borderline cases such as partially correct answers or cautious refusals. These findings confirm that GPT-4o provides a reliable and reproducible basis for large-scale automatic evaluation in \ours.

\begin{table}[!ht]
\centering
\resizebox{0.9\linewidth}{!}{
\begin{tabular}{lcc}
\toprule
\textbf{Model} & \textbf{Agreement (\%)} & \textbf{Cohen’s $\kappa$} \\
\midrule
\textsc{Keye-1.5-8B}     & 91.1 & 0.90 \\
\textsc{Ovis2-16B}       & 92.4 & 0.91 \\
\textsc{Mistral-3.1-24B} & 94.0 & 0.92 \\
\textsc{Claude-Opus-4}   & 93.0 & 0.92 \\
\midrule
\textbf{Average}         & \textbf{92.6} & \textbf{0.91} \\
\bottomrule
\end{tabular}
}
\caption{Human validation of GPT-4o judgments ($100$ samples per model, balanced across scenarios). We report raw agreement (\%) and Cohen’s $\kappa$. Averages are computed across the four models.}
\label{tab:human_validation}
\end{table}

\section{Performance Breakdown by sources}
While the main paper reports results averaged across sources, Table~\ref{tab:appendix_breakdown} provides a more detailed analysis, showing per-dataset performance for four representative LVLMs: \textsc{Keye-1.5-8B}, \textsc{Ovis2-16B}, \textsc{Mistral-Small-3.1-24B-Instruct}, and \textsc{Claude-Opus-4}. Three consistent patterns emerge.

First, performance varies substantially across datasets, revealing heterogeneous difficulty. \textsc{InfoSeek} is particularly challenging in the parametric setting, where accuracy remains near zero across all models, but improves markedly once gold evidence is available in the oracle scenario. In contrast, \textsc{WebQA} and \textsc{Encyclopedic-VQA} yield higher accuracy under both oracle and realistic conditions, while smaller datasets such as \textsc{ViQuAE} exhibit instability due to limited sample size, producing large metric fluctuations across models and scenarios.

Second, hallucination patterns closely mirror the availability of evidence. In the parametric scenario, hallucination is consistently high as models rely on internal memory. Once oracle passages are supplied, hallucination decreases across all sources, confirming that relevant grounding effectively constrains model generations. However, in adversarial settings, hallucination resurges sharply, demonstrating that carefully designed distractors remain highly effective at misleading LVLMs.

Finally, model ranking remains largely stable across sources. \textsc{Claude-Opus-4} and \textsc{Ovis2-16B} maintain higher accuracy and lower hallucination than \textsc{Keye-1.5-8B} and \textsc{Mistral-Small-3.1-24B}, though all systems exhibit vulnerabilities in adversarial contexts. Even for strong models, gains on \textsc{InfoSeek} under oracle grounding are modest, underscoring the intrinsic difficulty of encyclopedic reasoning.

Overall, this breakdown confirms that benchmark difficulty is not uniform: datasets differ both in absolute difficulty and in how models trade off accuracy, deflection, and hallucination. These variations highlight the importance of combining multiple KB-VQA sources to achieve a comprehensive and robust evaluation of retrieval-grounded reasoning.

\begin{table*}[!ht]
\centering
\resizebox{\textwidth}{!}{%
\begin{tabular}{ccccccccccccccccc}
\toprule
\multirow{2}{*}{\textbf{Model}} & 
\multirow{2}{*}{\textbf{Source}} & 
\multirow{2}{*}{\textbf{\# Samples}} &
\multicolumn{3}{c}{\textbf{Parametric}} &
\multicolumn{3}{c}{\textbf{Oracle}} &
\multicolumn{3}{c}{\textbf{Realistic}} &
\multicolumn{3}{c}{\textbf{Adversarial}} \\
\cmidrule(lr){4-6} \cmidrule(lr){7-9} \cmidrule(lr){10-12} \cmidrule(lr){13-15}
& & & Acc ↑ & Defl ↑ & Hall ↓ & Acc ↑ & Defl ↓ & Hall ↓ & Acc ↑ & Defl ↓ & Hall ↓ & Acc ↑ & Defl ↑ & Hall ↓ \\
\midrule
\multirow{7}{*}{Keye-1.5-8B} 
& Infoseek         & 1,646 & 1.6 & 17.1 & 81.3 & 56.7 & 11.7 & 31.6 & 47.1 & 10.8 & 42.1 & 0.9 & 38.4 & 60.7 \\
& WebQA            & 826 & 5.6 & 28.6 & 65.9 & 79.5 & 4.4 & 16.1 & 70.7 & 8.6 & 20.7 & 6.5 & 52.7 & 40.8 \\
& Encyclopedic-VQA & 185 & 3.8 & 27.0 & 69.2 & 77.3 & 11.9 & 10.8 & 73.5 & 9.7 & 16.8 & 2.2 & 66.5 & 31.4 \\
& MMDocRAG         & 58 & 0 & 31.0 & 69.0 & 69.0 & 0 & 31.0 & 60.3 & 0 & 39.7 & 15.5 & 12.1 & 72.4 \\
& MRAG-Bench       & 45 & 11.1 & 4.4 & 84.4 & 37.8 & 8.9 & 53.3 & 35.6 & 6.7 & 57.8 & 8.9 & 26.7 & 64.4 \\
& ViQuAE           & 15 & 6.7 & 33.3 & 60.0 & 46.7 & 40.0 & 13.3 & 53.3 & 20.0 & 26.7 & 6.6 & 46.7 & 46.7 \\
\rowcolor{ourlightgray}
& \textbf{Total}           & \textbf{2,775} & \textbf{3.1} & \textbf{21.4} & \textbf{75.5} & \textbf{64.8} & \textbf{9.4} & \textbf{25.8} & \textbf{56.0} & \textbf{9.8} & \textbf{34.2} & \textbf{3.1} & \textbf{43.8} & \textbf{53.1} \\
\cmidrule{1-15}
\multirow{7}{*}{Ovis2-16B} 
& Infoseek         & 1646 & 2.2 & 27.0 & 70.8 & 53.4 & 5.1 & 41.5 & 49.3 & 5.0 & 45.7 & 0.5 & 20.9 & 78.6 \\
& WebQA            & 826 & 7.6 & 19.1 & 73.3 & 86.1 & 1.0 & 12.9 & 82.6 & 0.2 & 17.2 & 8.7 & 24.9 & 66.4 \\
& Encyclopedic-VQA & 185 & 3.8 & 35.7 & 60.5 & 79.5 & 4.9 & 15.6 & 80.0 & 4.9 & 15.1 & 5.4 & 40.5 & 54.1 \\
& MMDocRAG         & 58 & 0 & 36.2 & 63.8 & 62.1 & 6.9 & 31.0 & 65.5 & 8.6 & 25.9 & 5.2 & 60.3 & 34.5 \\
& MRAG-Bench       & 45 & 11.1 & 20.0 & 68.9 & 11.1 & 24.4 & 64.5 & 13.3 & 8.9 & 77.8 & 11.1 & 11.3 & 75.6 \\
& ViQuAE           & 15 & 0 & 13.3 & 86.7 & 60.0 & 0 & 40 & 66.7 & 0 & 33.3 & 6.7 & 6.6 & 86.7 \\
\rowcolor{ourlightgray}
& \textbf{Total}           & \textbf{2,775} & \textbf{4.0} & \textbf{25.2} & \textbf{70.8} & \textbf{64.4} & \textbf{4.2} & \textbf{31.4} & \textbf{61.1} & \textbf{3.7} & \textbf{35.2} & \textbf{3.9} & \textbf{24.2} & \textbf{71.9} \\
\cmidrule{1-15}
\multirow{7}{*}{Mistral-Small-3.1} 
& Infoseek         & 1646 & 1.2 & 68.3 & 30.5 & 38.0 & 49.7 & 12.3 & 26.4 & 58.9 & 14.7 & 0.1 & 76.4 & 23.5 \\
& WebQA            & 826 & 5.1 & 73.1 & 21.8 & 48.1 & 45.6 & 6.3 & 42.6 & 52.3 & 5.1 & 0.8 & 96.6 & 2.6 \\
& Encyclopedic-VQA & 185 & 2.2 & 81.6 & 16.2 & 64.9 & 28.1 & 7.0 & 62.2 & 30.3 & 7.5 & 1.1 & 88.1 & 10.8 \\
& MMDocRAG         & 58 & 0 & 79.3 & 20.7 & 50.0 & 37.9 & 12.1 & 39.7 & 44.8 & 15.5 & 1.7 & 91.4 & 6.9 \\
& MRAG-Bench       & 45 & 8.9 & 66.7 & 24.4 & 6.7 & 66.7 & 26.6 & 2.2 & 91.1 & 6.7 & 4.4 & 91.1 & 4.5 \\
& ViQuAE           & 15 & 0 & 80.0 & 20.0 & 53.3 & 46.7 & 0 & 53.3 & 46.7 & 0 & 13.3 & 80.0 & 6.7 \\
\rowcolor{ourlightgray}
& \textbf{Total}           & \textbf{2,775} & \textbf{2.5} & \textbf{70.9} & \textbf{26.6} & \textbf{42.6} & \textbf{47.1} & \textbf{10.3} & \textbf{23.5} & \textbf{61.6} & \textbf{14.9} & \textbf{0.6} & \textbf{83.8} & \textbf{15.6} \\
\cmidrule{1-15}
\multirow{7}{*}{Claude-Opus-4} 
& Infoseek         & 1646 & 4.4 & 59.1 & 36.5 & 36.5 & 53.0 & 10.5 & 27.3 & 64.1 & 8.6 & 0.2 & 84.9 & 14.9 \\
& WebQA            & 826 & 11.3 & 71.9 & 16.8 & 71.2 & 21.1 & 7.7 & 63.8 & 28.5 & 7.7 & 0.8 & 94.9 & 4.3 \\
& Encyclopedic-VQA & 185 & 2.7 & 85.9 & 11.4 & 69.2 & 26.5 & 4.3 & 68.2 & 27.5 & 4.3 & 1.1 & 91.3 & 7.6 \\
& MMDocRAG         & 58 & 3.4 & 79.3 & 17.3 & 53.4 & 31.0 & 15.6 & 51.7 & 27.6 & 20.7 & 3.4 & 82.8 & 13.8 \\
& MRAG-Bench       & 45 & 20.0 & 48.9 & 31.1 & 11.1 & 82.2 & 6.7 & 11.1 & 71.1 & 17.8 & 6.7 & 91.1 & 2.2 \\
& ViQuAE           & 15 & 13.3 & 80.0 & 6.7 & 53.3 & 46.7 & 0 & 53.3 & 33.3 & 13.4 & 6.6 & 66.7 & 26.7 \\
\rowcolor{ourlightgray}
& \textbf{Total}           & \textbf{2,775} & \textbf{6.7} & \textbf{65.0} & \textbf{28.3} & \textbf{49.1} & \textbf{41.7} & \textbf{9.2} & \textbf{32.1} & \textbf{59.4} & \textbf{8.5} & \textbf{0.7} & \textbf{88.3} & \textbf{11.1} \\
\bottomrule
\end{tabular}%
}
\caption{Breakdown of performance across individual sources for four representative LVLMs. Results are reported in terms of accuracy, deflection, and hallucination under parametric, oracle, realistic, and adversarial settings. The table highlights heterogeneous dataset difficulty, systematic effects of external grounding on hallucination, and consistent relative model ranking across sources.}
\label{tab:appendix_breakdown}
\end{table*}

\section{Further Effect of Prompt Strictness}
\label{sec:appendix_severity}
Table~\ref{tab:appendix_severity} extends the main analysis of prompt severity by reporting results in the \textbf{parametric} and \textbf{oracle} scenarios. While the main paper highlighted clear trade-offs in the \textbf{realistic} and \textbf{adversarial} settings, these additional results reveal how refusal instructions behave in knowledge-free and perfectly grounded contexts. 
In the \textbf{parametric} case, where the target behavior is near-complete deflection, stronger prompts do encourage abstention. For example, \textsc{Claude-Opus-4} rises from $32.7\%$ to $97.8\%$ deflection when moving from no prompt to severe prompting, almost eliminating hallucination ($54.6\% \rightarrow 1.6\%$). A similar trend holds for \textsc{Mistral-Small-3.1}, which achieves $99.5\%$ deflection under severe prompting. These results confirm that explicit refusal instructions are highly effective when models truly lack knowledge.  
In the \textbf{oracle} setting, however, severe prompting leads to systematic over-deflection, even when gold knowledge is provided. For instance, \textsc{Ovis2-16B} accuracy falls from $65.0\%$ (soft) to $52.3\%$ (severe), while deflection rises sharply from $2.3\%$ to $25.1\%$. Similarly, \textsc{Claude-Opus-4} loses over $10$ points of accuracy under severe prompting, despite having access to perfect evidence. This pattern highlights a key limitation: while severity enforces abstention in knowledge-sparse settings, it also undermines performance when knowledge is available, effectively biasing models toward excessive caution.  
Overall, these results strengthen our main conclusion that prompt severity is a blunt tool: it can enforce deflection in parametric and adversarial contexts, but at the cost of suppressing accuracy in oracle and realistic ones. Reliable calibration therefore requires methods beyond simple refusal instructions, since current models cannot selectively adapt their abstention threshold to the evidence provided.

\begin{table}[!ht]
\centering
\resizebox{0.5\textwidth}{!}{%
\begin{tabular}{lccccccc}
\toprule
\multirow{2}{*}{\textbf{Model}} & 
\multirow{2}{*}{\textbf{Severity}} & 
\multicolumn{3}{c}{\textbf{Parametric}} &
\multicolumn{3}{c}{\textbf{Oracle}} \\
\cmidrule(lr){3-5} \cmidrule(lr){6-8}
& & Acc ↑ & Defl ↓ & Hall ↓ & Acc ↑ & Defl ↑ & Hall ↓ \\
\midrule
\multirow{4}{*}{Keye-1.5-8B} 
& None     & 4.6 & 4.5 & 90.9 & 68.0 & 2.9 & 29.1 \\
& Soft     & 4.1 & 11.4 & 84.5 & 65.8 & 6.9 & 27.3 \\
& Moderate & 3.1 & 21.4 & 75.5 & 64.8 & 9.4 & 25.8 \\
& Severe   & 0.6 & 82.3 & 17.1 & 37.5 & 47.0 & 15.5 \\
\cmidrule{1-8}
\multirow{4}{*}{Ovis2-16B} 
& None        & 5.3 & 9.6 & 85.1 & 63.0 & 1.2 & 35.8 \\
& Soft     & 4.0 & 24.8 & 71.2 & 65.0 & 2.3 & 32.7 \\
& Moderate & 4.0 & 25.2 & 70.8 & 64.4 & 4.2 & 31.4 \\
& Severe   & 0.7 & 89.5 & 9.8 & 52.3 & 25.1 & 22.6 \\
\cmidrule{1-8}
\multirow{4}{*}{Mistral-Small-3.1} 
& None        & 6.4 & 33.3 & 60.3 & 66.2 & 14.1 & 19.7 \\
& Soft     & 4.5 & 48.6 & 46.9 & 61.0 & 22.5 & 16.5 \\
& Moderate & 2.5 & 70.9 & 26.6 & 42.6 & 47.1 & 10.3 \\
& Severe   & 0.1 & 99.5 & 0.4 & 13.3 & 85.8 & 0.9 \\
\cmidrule{1-8}
\multirow{4}{*}{Claude-Opus-4} 
& None        & 12.7 & 32.7 & 54.6 & 52.4 & 37.0 & 10.6 \\
& Soft     & 9.1 & 49.0 & 41.9 & 51.6 & 37.0 & 11.4 \\
& Moderate & 6.7 & 65.0 & 28.3 & 49.1 & 41.7 & 9.2 \\
& Severe   & 0.6 & 97.8 & 1.6 & 40.0 & 53.1 & 6.9 \\
\bottomrule
\end{tabular}
}
\caption{Deflection prompt strictness effects in the \textbf{Parametric} and \textbf{Oracle} scenarios across four LVLMs. Increasing strictness successfully raises deflection rates and reduces hallucination in strong models, but comes at the cost of significant accuracy degradation, revealing fundamental trade-offs in reliability tuning.}
\label{tab:appendix_severity}
\end{table}

\section{Implementation Details}
\label{sec:implementation}

We employ a CLIP-based encoder (\textit{i.e.}, EVA-CLIP-18B~\cite{sun2024eva}) to encode either textual metadata or visual content into dense vectors $w_i \in \mathbb{R}^d$, which are stored in a similarity-search index $W = \{w_i\}_i$. Retrieval is carried out via cosine similarity between query embeddings and indexed features. For each benchmark, we systematically explore both retrieval modalities: (i) \textbf{Textual}, where an input image is mapped to Wikipedia text (either titles only (T) or titles plus summaries (T+S)); and (ii) \textbf{Visual}, where similarity is computed directly between images (I2I). The retrieval modality with highest performance on each dataset is selected. 
For \textbf{InfoSeek}, we adopt image-to-text retrieval (I2T). The knowledge base is built from the provided English Wikipedia dump, containing roughly 6M entities. After retrieving the top-10 candidate pages, we further segment each page into 250-token passages with 32-token overlap using \textsc{Contriever}~\cite{izacardunsupervised}, ensuring well-formed sentence boundaries and more targeted retrieval.  
For \textbf{Encyclopedic-VQA}, we instead rely on image-to-image retrieval (I2I). The knowledge base here is the dataset-provided collection of Wikipedia pages (about 2M entries), which differs from the full 6M-page dump used for InfoSeek. Query images are compared against the first image of each page, and the top-5 retrieved pages are subsequently segmented into 250-token passages with 32-token overlap using the same \textsc{Contriever}-based procedure.  
For \textbf{MRAG-Bench}, which supplies its own knowledge base, we construct an index from the provided images and perform I2I retrieval of top-$20$ examples, excluding those overlapping with gold evidence. \textbf{ViQuAE} follows the same text-based strategy as InfoSeek (I2T with page segmentation). In contrast, \textbf{WebQA} and \textbf{MMDocRAG} require no external retrieval, as positive and negative candidates are already provided.  
Notably, in principle the retrieval process could be reduced to a simple overlap filtering step with gold evidence. However, we design a more complex pipeline to prioritize the generation of challenging and plausible negative contexts, rather than maximizing retrieval precision. This choice ensures that the benchmarks serve their primary purpose: evaluating model robustness under difficult distractor conditions.

\section{Evaluation Prompts}
Figure~\ref{fig:appendix_system_prompt} illustrates the \textbf{system prompt} used to constrain models to rely only on the provided context. All evaluations were conducted with the \textbf{moderate prompt} (Figure~\ref{fig:appendix_moderate_prompt}), which instructs models to answer only when sufficiently confident. For the ablation experiments on prompt strictness (Tables~\ref{tab:table_severity} and~\ref{tab:appendix_severity}), we additionally considered the \textbf{soft} and \textbf{severe} variants (Figures~\ref{fig:appendix_soft_prompt} and~\ref{fig:appendix_severe_prompt}), corresponding to looser and stricter abstention requirements, respectively. These variants allow us to analyze how LVLMs trade off coverage and reliability under different answering conditions.
\label{sec:appendix_prompts}

\section{Qualitative Analysis and Failure Modes}
\label{sec:appendix_qualitative}
\begin{figure*}[t]
\centering
\begin{tcolorbox}[
  enhanced,
  width=\textwidth,
  colback=white,
  colframe=gray!30!black,
  boxrule=0.45pt,
  arc=4pt,
  left=10pt, right=10pt, top=10pt, bottom=10pt,
  shadow={0.8ex}{-0.8ex}{0.6ex}{gray!35!white},
]

\begin{minipage}[c]{0.28\linewidth}
  \includegraphics[width=\linewidth]{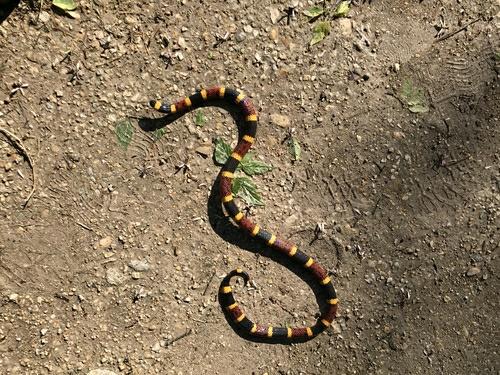}
\end{minipage}\hfill
\begin{minipage}[c]{0.68\linewidth}
  \begin{flushright}
\begin{tcolorbox}[colback=gray!10, colframe=gray!40, 
  boxrule=0.4pt, arc=2pt, left=6pt, right=6pt, top=2pt, bottom=2pt]
  \centering \textbf{Source:} \textsc{Encyclopedic-VQA}
\end{tcolorbox}
\end{flushright}
  \vspace{4pt}

  \small
  \textbf{Context:} The primary diet of \textit{M.~tener} consists of other snakes, primarily earth snakes, and other small fossorial species. It is cannibalistic. It also occasionally eats small lizards, but the consumption of rodents by coral snakes is rare.
\end{minipage}

\vspace{6pt}

\begin{minipage}{0.95\linewidth}
  \small
  \textbf{Question:} What does this reptile eat?\\[2pt]
  \textbf{GT Answer:} Earth snakes, fossorial species.\\[4pt]

  \textbf{OVIS2-16B:} This reptile, which appears to be a coral snake, primarily eats other snakes, particularly earth snakes and other small fossorial species. It is also known to be cannibalistic and occasionally consumes small lizards. The consumption of rodents by coral snakes is rare.  
  \hfill \textsc{Judge: \textcolor{CorrectColor}{Correct}}\\[4pt]

  \textbf{MISTRAL-3.1-24B:} I cannot provide an answer based on the given information because the context provided refers to a different species, \textit{M.~tener}, and does not specify the species of the snake in the image.  
  \judge{DeflectColor}{Not\_Attempted}\\[4pt]

  \textbf{CLAUDE-OPUS-4:} Based on the retrieved information provided, this reptile (which appears to be a coral snake based on its distinctive banding pattern) primarily eats other snakes, especially earth snakes and other small fossorial species. It is also cannibalistic, meaning it will eat other coral snakes. Additionally, it occasionally consumes small lizards, though rodent consumption is rare.  
  \hfill \textsc{Judge: \textcolor{CorrectColor}{Correct}}
\end{minipage}

\end{tcolorbox}

\caption{Example from \textsc{Encyclopedic-VQA} (\textbf{Oracle}). All models are given gold textual evidence about the coral snake’s diet. \textsc{Ovis2-16B} and \textsc{Claude-Opus-4} ground their answers correctly, while \textsc{Mistral-3.1-24B} deflects despite sufficient information, illustrating conservative behavior under clear evidence.}
\label{fig:qual_1}
\end{figure*}

\begin{figure*}[t]
\centering
\begin{tcolorbox}[
  enhanced,
  width=\textwidth,
  colback=white,
  colframe=gray!30!black,
  boxrule=0.45pt,
  arc=4pt,
  left=10pt, right=10pt, top=10pt, bottom=10pt,
  shadow={0.8ex}{-0.8ex}{0.6ex}{gray!35!white},
]

\begin{minipage}[c]{0.28\linewidth}
  \includegraphics[width=\linewidth]{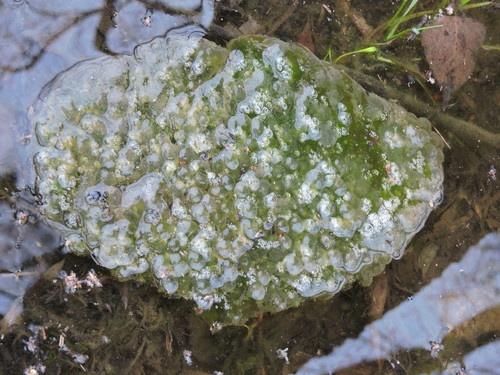}
\end{minipage}\hfill
\begin{minipage}[c]{0.68\linewidth}
  \begin{flushright}
  \begin{tcolorbox}[colback=gray!10, colframe=gray!40, 
    boxrule=0.4pt, arc=2pt, left=6pt, right=6pt, top=2pt, bottom=2pt]
    \centering \textbf{Source:} \textsc{Encyclopedic-VQA}
  \end{tcolorbox}
  \end{flushright}
  \vspace{4pt}
  \small
  \textbf{Context:} The agile frog prefers light deciduous mixed forests with plentiful water. The open land around a forest is often also populated, as long as it is connected to the forest by shrubs. In dry, warm forests, this species often also lives far away from the water. Of the three Middle European \textit{Rana} species, this frog likes warmth and dryness the most.
\end{minipage}

\vspace{6pt}

\begin{minipage}{0.95\linewidth}
  \small
  \textbf{Question:} What is the habitat of this animal?\\[2pt]
  \textbf{GT Answer:} Mixed forests with plentiful water.\\[4pt]

  \textbf{OVIS2-16B:} The habitat of the agile frog includes light deciduous mixed forests with plentiful water, open land connected to the forest by shrubs, and dry, warm forests.  
  \hfill \textsc{Judge: \textcolor{CorrectColor}{Correct}}\\[4pt]

  \textbf{MISTRAL-3.1-24B:} I cannot confidently determine the habitat of the organism in the image based on the provided information. The retrieved context discusses the habitat preferences of the agile frog, but there is no indication that the organism in the image is an agile frog. Therefore, I cannot provide an accurate answer.  
  \judge{DeflectColor}{Not\_Attempted}\\[4pt]

  \textbf{CLAUDE-OPUS-4:} I cannot provide information about the habitat of the animal in the image based on the retrieved context provided. The context discusses the habitat preferences of the agile frog, but I cannot confirm whether the animal in the image is an agile frog or another species. Without being able to definitively identify the animal in the image as an agile frog, I cannot reliably apply the habitat information from the context.  
  \judge{DeflectColor}{Not\_Attempted}
\end{minipage}

\end{tcolorbox}

\caption{Example from \textsc{Encyclopedic-VQA} (\textbf{Oracle}). Gold text specifies the habitat; \textsc{Ovis2-16B} answers correctly while \textsc{Mistral-3.1-24B} and \textsc{Claude-Opus-4} over-deflect under gold evidence.}
\label{fig:qual_2}
\end{figure*}

\begin{figure*}[t]
\centering
\begin{tcolorbox}[
  enhanced,
  width=\textwidth,
  colback=white,
  colframe=gray!30!black,
  boxrule=0.45pt,
  arc=4pt,
  left=10pt, right=10pt, top=10pt, bottom=10pt,
  shadow={0.8ex}{-0.8ex}{0.6ex}{gray!35!white},
]

\begin{minipage}[c]{0.28\linewidth}
  \includegraphics[width=\linewidth]{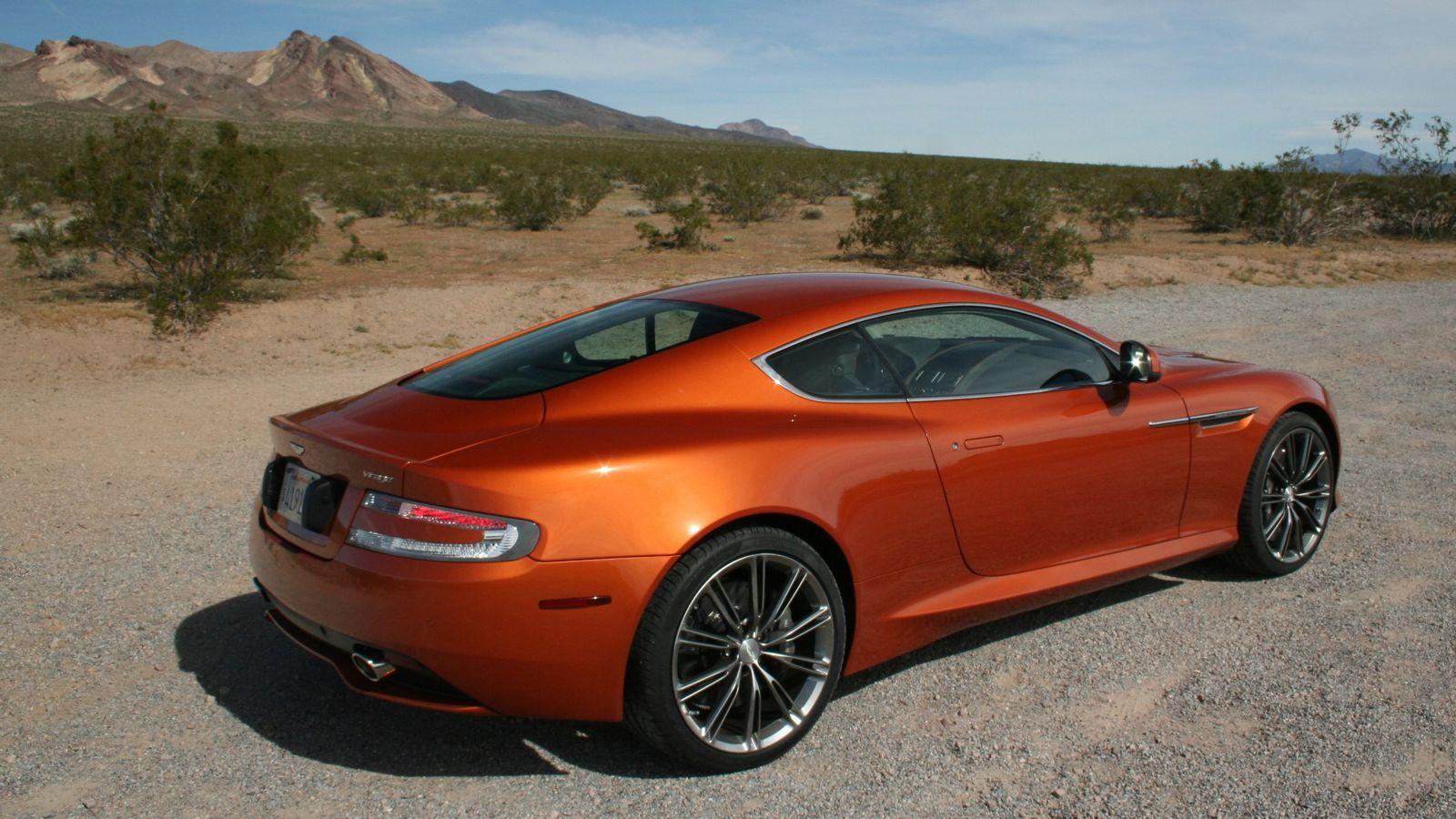}
\end{minipage}\hfill
\begin{minipage}[c]{0.68\linewidth}
  \begin{flushright}
  \begin{tcolorbox}[colback=gray!10, colframe=gray!40, 
    boxrule=0.4pt, arc=2pt, left=6pt, right=6pt, top=2pt, bottom=2pt]
    \centering \textbf{Source:} \textsc{MRAG-Bench}
  \end{tcolorbox}
  \end{flushright}

  \vspace{4pt}
  \raggedright \small \textbf{Context:}\\[4pt]
  \centering
  \includegraphics[width=0.23\linewidth]{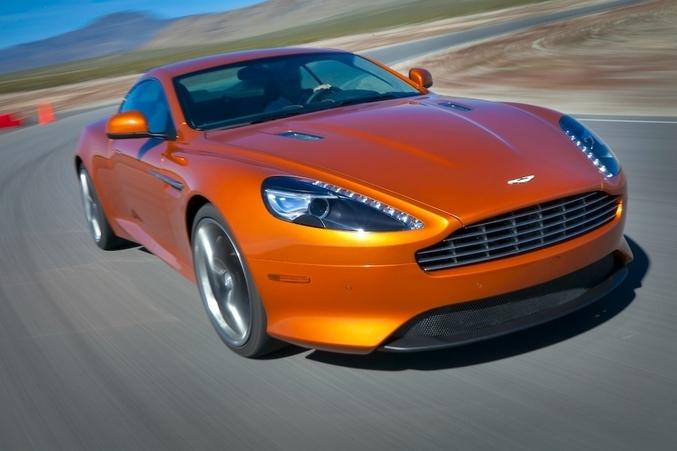}
  \includegraphics[width=0.23\linewidth]{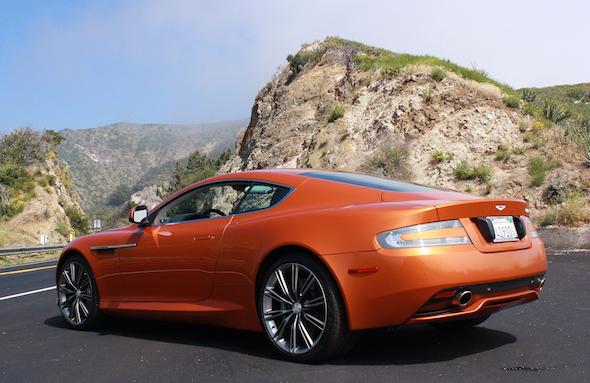}
  \includegraphics[width=0.23\linewidth]{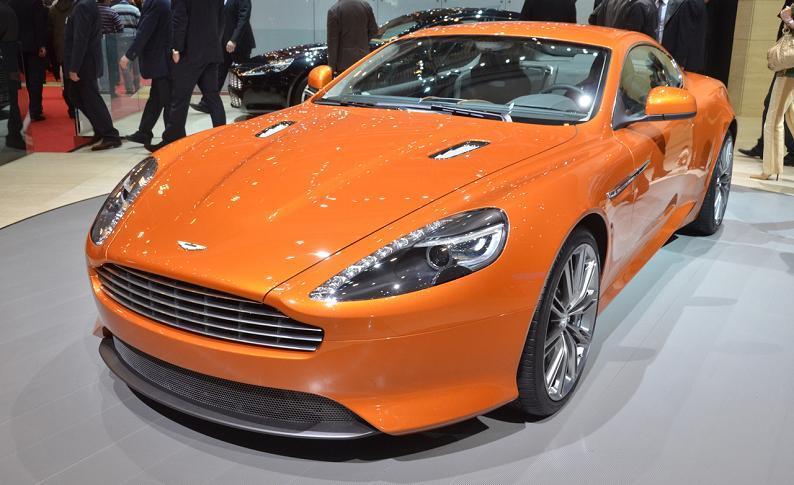}
  \includegraphics[width=0.23\linewidth]{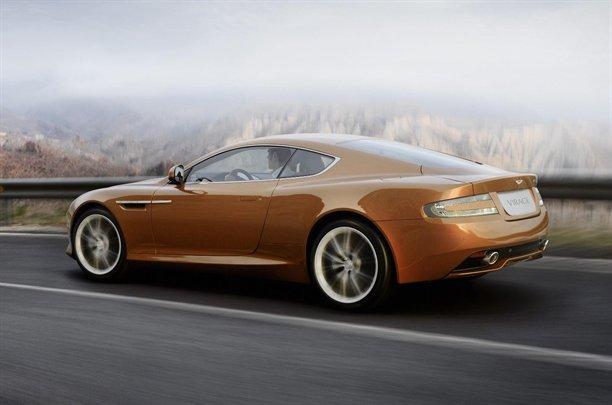}
\end{minipage}

\vspace{6pt}

\begin{minipage}{0.95\linewidth}
  \small
  \textbf{Question:} Can you identify the particular model of this car?\\[2pt]
  \textbf{GT Answer:} Aston Martin Virage Coupe 2012.\\[4pt]

  \textbf{OVIS2-16B:} Based on the visual characteristics such as the sleek design, the shape of the taillights, and the overall silhouette, the car appears to be an Aston Martin DB9. This identification is made by comparing the visible features with known models from Aston Martin's lineup. However, without a clear view of specific badges or additional identifying marks, this identification cannot be guaranteed with absolute certainty. \hfill \textsc{Judge: \textcolor{HallucinateColor}{Incorrect}}\\[4pt]

  \textbf{MISTRAL-3.1-24B:} I cannot definitively identify the exact model of the car from the image alone. However, based on the design and features visible, it appears to be a high-performance sports car, possibly from a luxury brand. For precise identification, additional details or a clearer view of specific badges or emblems would be necessary. \hfill \textsc{Judge: \textcolor{DeflectColor}{Not\_Attempted}}\\[4pt]

  \textbf{CLAUDE-OPUS-4:} Based on the retrieved images provided, I cannot definitively identify the particular model of this car. While the retrieved images show similar orange/copper-colored luxury sports cars with comparable styling, no text or captions are provided with these images that explicitly identify the specific model. Without additional contextual information such as model names, badges, or descriptive text in the retrieved content, I cannot provide a reliable answer about the particular model of this vehicle. \hfill \textsc{Judge: \textcolor{DeflectColor}{Not\_Attempted}}
\end{minipage}

\end{tcolorbox}

\caption{Example from \textsc{MRAG-Bench} (\textbf{Oracle}, visual gold). \textsc{Ovis2-16B} hallucinates a wrong car model; \textsc{Mistral-3.1-24B} and \textsc{Claude-Opus-4} abstain. A grounding failure with gold images.}
\label{fig:qual_3}
\end{figure*}

\begin{figure*}[t]
\centering
\begin{tcolorbox}[
  enhanced,
  width=\textwidth,
  colback=white,
  colframe=gray!30!black,
  boxrule=0.45pt,
  arc=4pt,
  left=10pt, right=10pt, top=10pt, bottom=10pt,
  shadow={0.8ex}{-0.8ex}{0.6ex}{gray!35!white},
]

\begin{minipage}[c]{0.28\linewidth}
  \includegraphics[width=\linewidth]{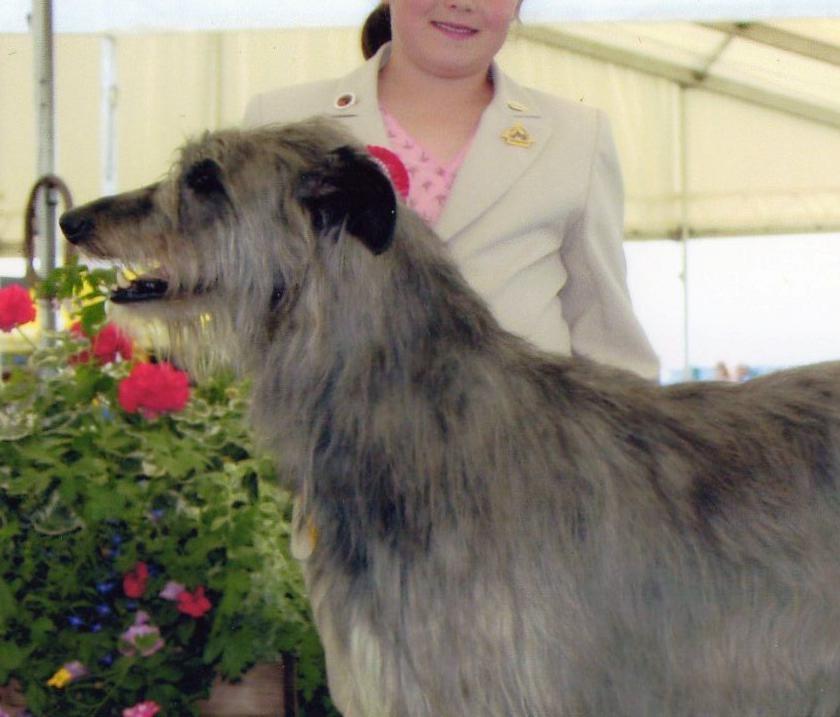}
\end{minipage}\hfill
\begin{minipage}[c]{0.68\linewidth}
  \begin{flushright}
  \begin{tcolorbox}[colback=gray!10, colframe=gray!40,
    boxrule=0.4pt, arc=2pt, left=6pt, right=6pt, top=2pt, bottom=2pt]
    \centering \textbf{Source:} \textsc{MRAG-Bench}
  \end{tcolorbox}
  \end{flushright}

  \vspace{4pt}
  \raggedright \small \textbf{Context:}\\[4pt]
  \centering
  \includegraphics[width=0.23\linewidth]{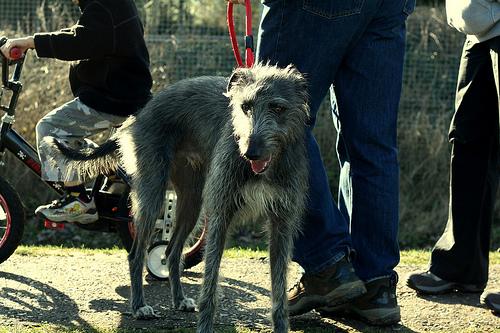}
  \includegraphics[width=0.23\linewidth]{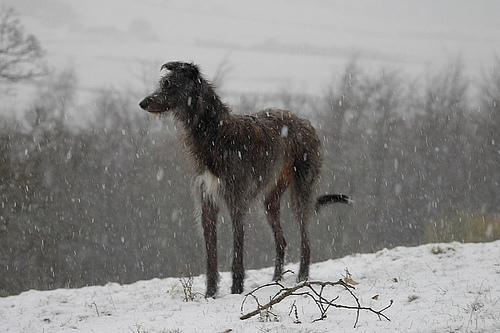}
  \includegraphics[width=0.23\linewidth]{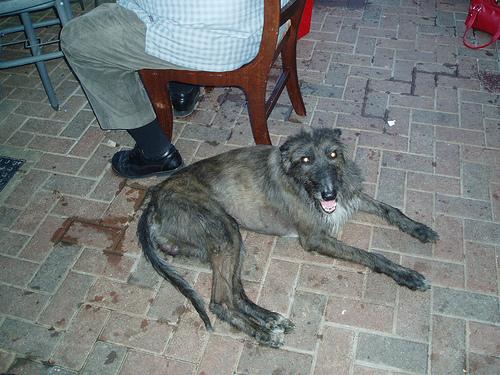}
  \includegraphics[width=0.23\linewidth]{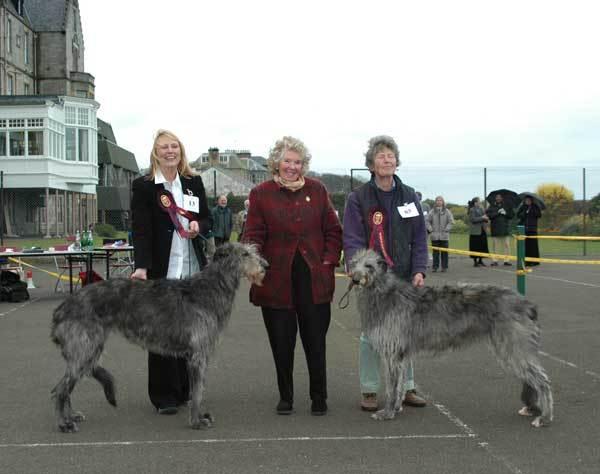}
\end{minipage}

\vspace{6pt}

\begin{minipage}{0.95\linewidth}
  \small
  \textbf{Question:} Can you identify what creature this is?\\[2pt]
  \textbf{GT Answer:} Scottish Deerhound.\\[4pt]

  \textbf{OVIS2-16B:} The creature in the image is a dog, specifically a breed that resembles an Afghan Hound, known for its long, silky coat and distinctive appearance. This identification is based on the visible physical characteristics of the dog, such as its long fur, slender build, and the shape of its ears and muzzle. The setting suggests that the dog may be participating in a dog show, which is a common context for displaying purebred dogs. \hfill \textsc{Judge: \textcolor{HallucinateColor}{Incorrect}}\\[4pt]

  \textbf{MISTRAL-3.1-24B:} The creature in the image is a dog. Specifically, it appears to be a Deerhound, which is a breed known for its shaggy coat and tall, slender build. \hfill \textsc{Judge: \textcolor{CorrectColor}{Correct}}\\[4pt]

  \textbf{CLAUDE-OPUS-4:} Based on the retrieved images provided, I can see several dogs that appear to be similar to the creature in the input image. The retrieved images show various dogs with long, shaggy, grey coats that match the characteristics of the animal in the question image. However, without any text or specific breed information provided in the retrieved context, I cannot identify the specific breed of dog this is. I can only confirm based on the visual similarity to the retrieved images that this appears to be a dog with a distinctive long, grey, shaggy coat. 
  \judge{DeflectColor}{Not\_Attempted}\\
\end{minipage}

\end{tcolorbox}

\caption{Example from \textsc{MRAG-Bench} (\textbf{Oracle}, visual gold). \textsc{Mistral-3.1-24B} identifies the breed correctly; \textsc{Ovis2-16B} hallucinates; \textsc{Claude-Opus-4} over-deflects. Calibration varies despite gold images.}
\label{fig:qual_4}
\end{figure*}

\begin{figure*}[t]
\centering
\begin{tcolorbox}[
  enhanced,
  width=\textwidth,
  colback=white,
  colframe=gray!30!black,
  boxrule=0.45pt,
  arc=4pt,
  left=10pt, right=10pt, top=10pt, bottom=10pt,
  shadow={0.8ex}{-0.8ex}{0.6ex}{gray!35!white},
]

\centerline{%
  \tcbox[colback=gray!10, colframe=gray!40, boxrule=0.4pt, arc=2pt,
         left=6pt, right=6pt, top=2pt, bottom=2pt]{%
    \textbf{Source:} \textsc{MMDocRAG}%
  }%
}

\vspace{4pt}

\begin{minipage}[c]{0.35\linewidth}
  \raggedright \small \textbf{Context (Images):}\\[2pt]
  \includegraphics[width=0.8\linewidth]{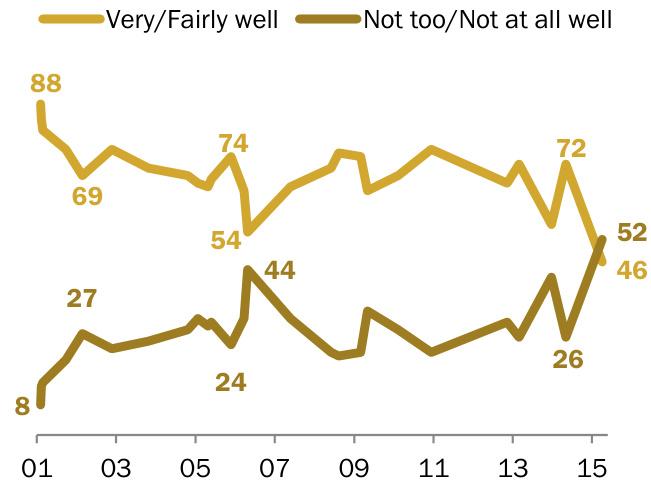}\\[2pt]
  \includegraphics[width=0.8\linewidth]{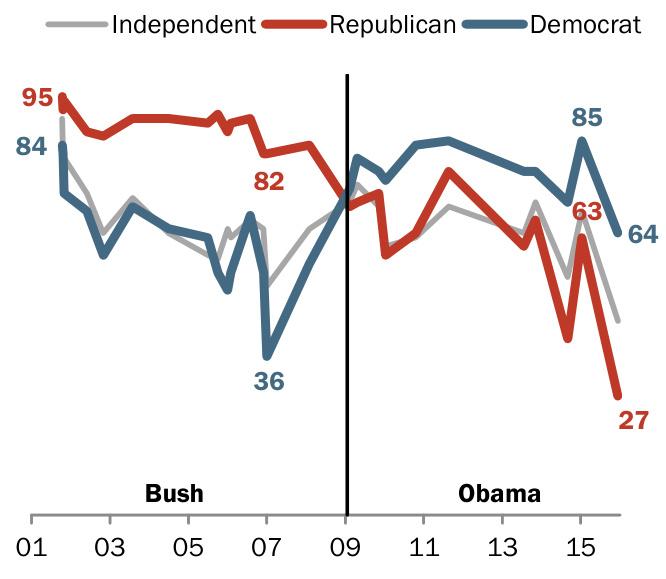}
\end{minipage}\hfill
\begin{minipage}[c]{0.65\linewidth}
  \raggedright \small \textbf{Context (Text):}\\[2pt]
  1. The latest national survey by Pew Research Center, conducted Dec. 8–13 among 1,500 adults, finds that since the start of this year, the share of Americans who say the government is doing well in reducing the threat of terrorism has fallen by 26 percentage points – from $72\%$ to $46\%$.\\
  \vspace{1em}
  2. Compared to early 2015, assessments of government efforts to combat terrorism are more negative across the political spectrum. Democrats are now the only partisan group in which a majority $(64\%)$ say the government is doing at least fairly well (down from $85\%$ in January). Independents’ positive ratings have dropped 25 points, from $69\%$ to $44\%$. And just $27\%$ of Republicans now say the government is doing very or fairly well reducing the terrorist threat, down from $63\%$ at the beginning of the year.
\end{minipage}

\vspace{6pt}

\begin{minipage}{0.95\linewidth}
  \small
  \textbf{Question:} In 2015, how many percentage of surveyed adults, Republicans, and Democrats believed that the government was doing very/fairly well in reducing the threat of terrorism? Please write the answer in list format, \textit{e.g.}, [3, 2]\\[2pt]
  \textbf{GT Answer:} [``46'', ``27'', ``64''].\\[4pt]

  \textbf{OVIS2-16B:} [``46\%'', ``27\%'', ``64\%''] \hfill \textsc{Judge: \textcolor{CorrectColor}{Correct}}\\[4pt]

  \textbf{MISTRAL-3.1-24B:} I cannot provide an answer based on the given information because the data provided is from December 2015, not from earlier in the year. \hfill \textsc{Judge: \textcolor{DeflectColor}{Not\_Attempted}} \\[4pt]

  \textbf{CLAUDE-OPUS-4:} Based on the retrieved information, I can provide the percentages for early 2015 (January): 72\%, 63\%, and 85\%. Therefore, the answer in list format is: [``72'', ``63'', ``85'']. \hfill \textsc{Judge: \textcolor{HallucinateColor}{Incorrect}}
\end{minipage}

\end{tcolorbox}

\caption{Example from \textsc{MMDocRAG} (\textbf{Oracle}). Gold text supports exact numeric extraction: \textsc{Ovis2-16B} is correct; \textsc{Mistral-3.1-24B} over-deflects; \textsc{Claude-Opus-4} hallucinates outdated figures.}
\label{fig:qual_5}
\end{figure*}

\begin{figure*}[t]
\centering
\begin{tcolorbox}[
  enhanced,
  width=\textwidth,
  colback=white,
  colframe=gray!30!black,
  boxrule=0.45pt,
  arc=4pt,
  left=10pt, right=10pt, top=10pt, bottom=10pt,
  shadow={0.8ex}{-0.8ex}{0.6ex}{gray!35!white},
]

\centerline{%
  \tcbox[colback=gray!10, colframe=gray!40, boxrule=0.4pt, arc=2pt,
         left=6pt, right=6pt, top=2pt, bottom=2pt]{%
    \textbf{Source:} \textsc{WebQA}%
  }%
}

\vspace{4pt}

\begin{minipage}[c]{0.25\linewidth}
  \raggedright \small \textbf{Context (Images):}\\[2pt]
  \includegraphics[width=0.8\linewidth]{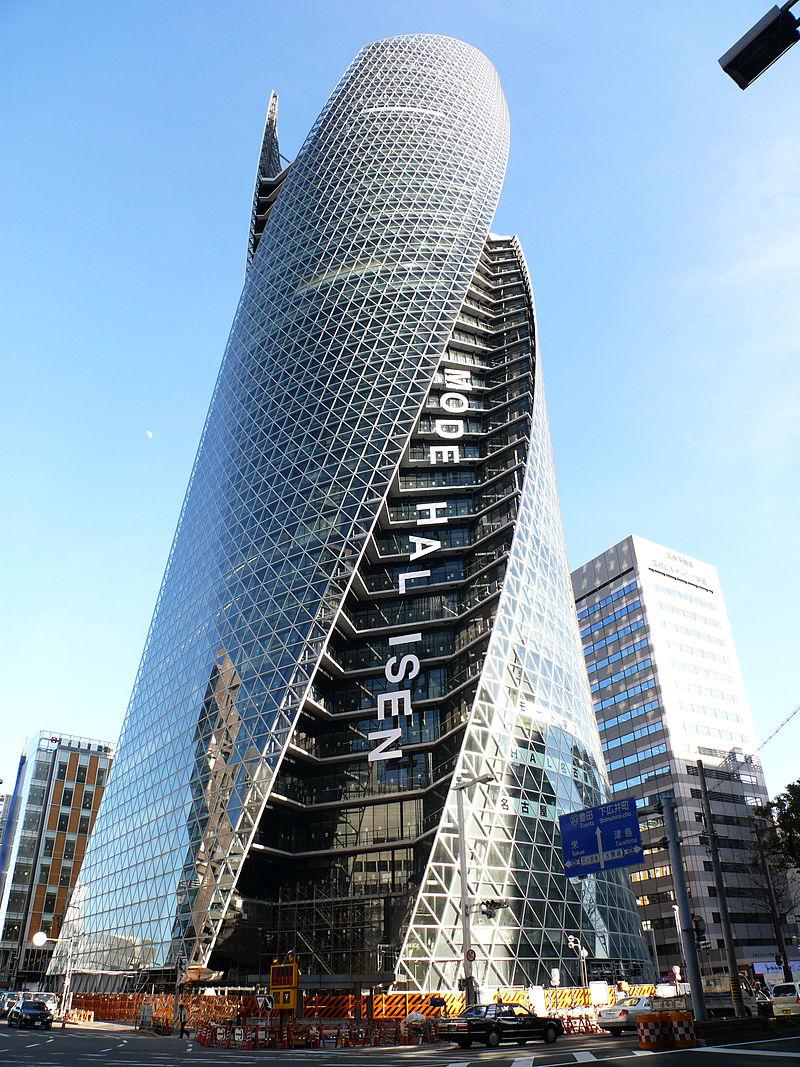}
\end{minipage}\hfill
\begin{minipage}[c]{0.75\linewidth}
  \raggedright \small \textbf{Context (Text):}\\[2pt]
  \textsc{NO CONTEXT} \\[20pt]
  \small
  \textbf{Question:} How many letters are found on the outside of the Nagoya Mode Academy Spiral Towers? \\[2pt]
  \textbf{GT Answer:} 11 letters are found on the outside of the Nagoya Mode Academy Spiral Towers.\\[4pt]
\end{minipage}

\vspace{6pt}

\begin{minipage}{0.95\linewidth}
  \small
  \textbf{OVIS2-16B:} There are 12 letters found on the outside of the Nagoya Mode Academy Spiral Towers. 
  \judge{HallucinateColor}{Incorrect}\\[4pt]

  \textbf{MISTRAL-3.1-24B:} I am unable to determine the number of letters on the outside of the Nagoya Mode Academy Spiral Towers based on the provided information. \hfill \textsc{Judge: \textcolor{DeflectColor}{Not\_Attempted}} \\[4pt]

  \textbf{CLAUDE-OPUS-4:} Based on the retrieved image, I can see letters displayed vertically on the outside of the spiral tower building. The visible letters spell out “MODE GAKUEN.” Counting these letters, there are 10. \judge{HallucinateColor}{Incorrect}\\
\end{minipage}

\end{tcolorbox}

\caption{Example from \textsc{WebQA} (\textbf{Oracle}, visual gold). Counting letters from the gold image: \textsc{Ovis2-16B} and \textsc{Claude-Opus-4} hallucinate, \textsc{Mistral-3.1-24B} over-deflects.}
\label{fig:qual_6}
\end{figure*}

\begin{figure*}[t]
\centering
\begin{tcolorbox}[
  enhanced,
  width=\textwidth,
  colback=white,
  colframe=gray!30!black,
  boxrule=0.45pt,
  arc=4pt,
  left=10pt, right=10pt, top=10pt, bottom=10pt,
  shadow={0.8ex}{-0.8ex}{0.6ex}{gray!35!white},
]

\centerline{%
  \tcbox[colback=gray!10, colframe=gray!40, boxrule=0.4pt, arc=2pt,
         left=6pt, right=6pt, top=2pt, bottom=2pt]{%
    \textbf{Source:} \textsc{WebQA}%
  }%
}

\vspace{4pt}

\begin{minipage}[c]{0.8\linewidth}
  \raggedright \small \textbf{Context (Images):}\\[2pt]
  \includegraphics[width=0.35\linewidth]{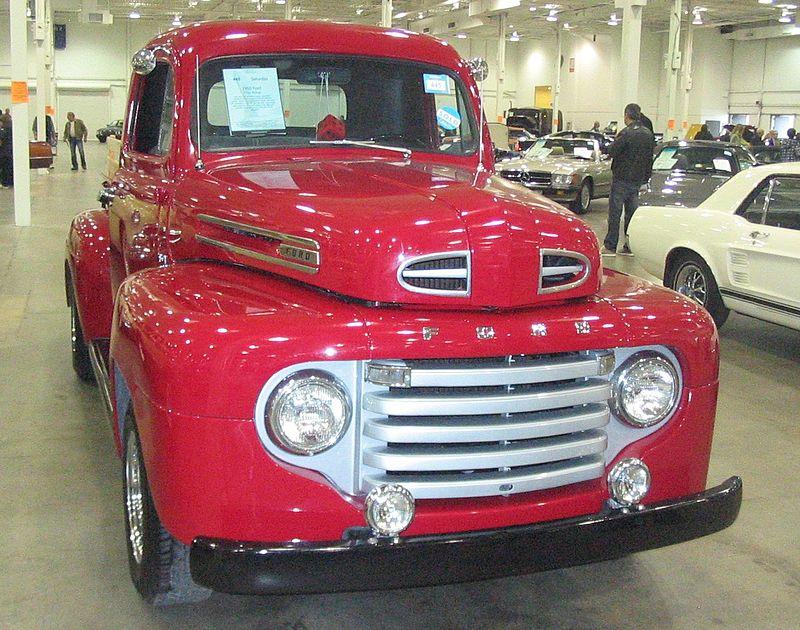}
  \includegraphics[width=0.37\linewidth]{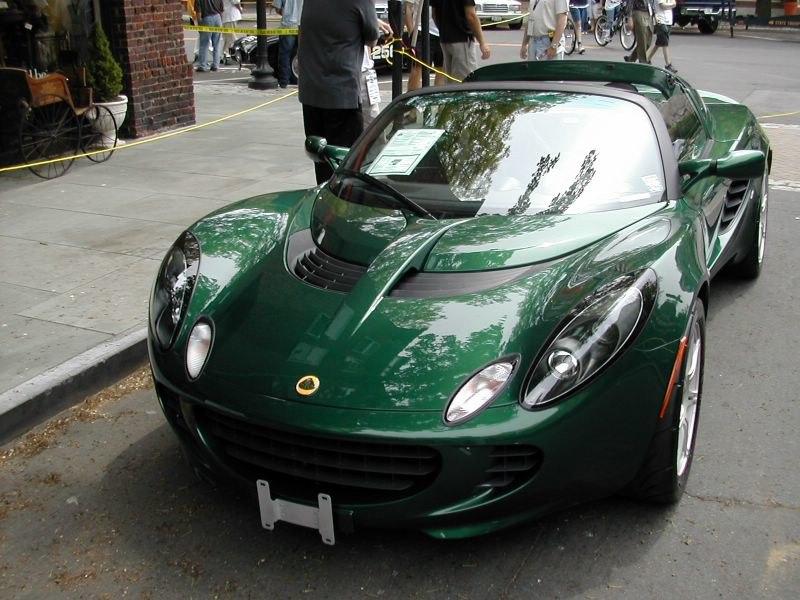}
\end{minipage}\hfill
\begin{minipage}[c]{0.2\linewidth}
  \raggedright \small \textbf{Context (Text):}\\[2pt]
  \textsc{NO CONTEXT}
\end{minipage}

\vspace{6pt}

\begin{minipage}{0.95\linewidth}
  \small
  \textbf{Question:} Which is taller: the '50 Ford F-Series or Lotus Elise?\\[2pt]
  \textbf{GT Answer:} The ’50 Ford F-Series is taller.\\[4pt]

  \textbf{OVIS2-16B:} Based on the images provided, the ’50 Ford F-Series appears to be taller than the Lotus Elise. The Ford F-Series has a higher roofline and overall height compared to the Lotus Elise, which has a lower profile typical of sports cars. \hfill \textsc{Judge: \textcolor{CorrectColor}{Correct}}\\[4pt]

  \textbf{MISTRAL-3.1-24B:} I cannot determine which vehicle is taller based solely on the images provided.
  \judge{DeflectColor}{Not\_Attempted}\\[4pt]

  \textbf{CLAUDE-OPUS-4:} Based on the provided images, I cannot determine which vehicle is taller. While I can see that the Ford F-Series is a truck with a higher roofline and the Lotus Elise is a low-slung sports car, the lack of scale or shared reference prevents reliable comparison. \hfill \textsc{Judge: \textcolor{DeflectColor}{Not\_Attempted}}
\end{minipage}

\end{tcolorbox}

\caption{Example from \textsc{WebQA} (\textbf{Oracle}, visual gold). With two gold images, \textsc{Ovis2-16B} correctly infers the taller object; \textsc{Mistral-3.1-24B} and \textsc{Claude-Opus-4} over-deflect.}
\label{fig:qual_7}
\end{figure*}

Figures~\ref{fig:qual_1}--\ref{fig:qual_7} show representative \textbf{Oracle} cases (gold evidence provided). The examples surface three recurring behaviors under perfect evidence: (i) \emph{correct grounding} (\textit{e.g.}, \textsc{Ovis2-16B}, \textsc{Claude-Opus-4} in Figures~\ref{fig:qual_1}, \ref{fig:qual_5}); (ii) \emph{over-deflection} despite sufficient evidence (\textit{e.g.}, \textsc{Mistral-3.1-24B} in Figures~\ref{fig:qual_1}, \ref{fig:qual_2}, and \ref{fig:qual_6}); and (iii) \emph{hallucination with gold context} (\textit{e.g.}, \textsc{Ovis2-16B} in Figures~\ref{fig:qual_3}, \ref{fig:qual_4}, and \textsc{Claude-Opus-4} in Figure~\ref{fig:qual_5}, \ref{fig:qual_6}). 

\textbf{Textual gold} (Figures~\ref{fig:qual_1}, \ref{fig:qual_2}, \ref{fig:qual_5}) often enables precise answers, yet some models still refuse (over-deflect) or produce unsupported numbers. \textbf{Visual gold} (Figures~\ref{fig:qual_3}, \ref{fig:qual_4}, \ref{fig:qual_6}, \ref{fig:qual_7}) exposes calibration gaps: models either misidentify fine-grained categories or abstain even when the image suffices. These cases confirm that remaining errors are not due to retrieval quality but to \emph{evidence utilization}: models must better distinguish when gold evidence is adequate to answer versus when abstention is warranted.

\begin{figure*}[htbp]
    \begin{tcolorbox}[colback=gray!5, colframe=gray!40!black, title=System Prompt, width=\textwidth]
    \ttfamily
    You are a helpful assistant in a Retrieval-Augmented Generation (RAG) scenario. \\You may receive:\\
    - Textual information from retrieved documents\\
    - Visual information from images\\
    - Multimodal content combining both text and images
    \end{tcolorbox}
    \caption{System prompt used in our RAG experiments showing the initial instructions given to the model.}
    \label{fig:appendix_system_prompt}
\end{figure*}

\begin{figure*}[htbp]
    \begin{tcolorbox}[colback=green!5, colframe=green!40!black, title=SOFT Uncertainty Level Prompt, width=\textwidth]
    \ttfamily
    Answer the question based on the available information. You may abstain if necessary. Answer:
    \end{tcolorbox}
    \caption{Soft prompt variant encouraging models to answer based on available information with optional deflection when necessary.}
    \label{fig:appendix_soft_prompt}
\end{figure*}

\begin{figure*}[htbp]
    \begin{tcolorbox}[colback=orange!5, colframe=orange!40!black, title=MODERATE Uncertainty Level Prompt, width=\textwidth]
    \ttfamily
    Please provide an answer only if you feel confident in its accuracy. If you are uncertain or the available information is insufficient to provide a reliable answer, please abstain rather than guessing. Answer:
    \end{tcolorbox}
    \caption{Moderate prompt variant instructing models to answer only when sufficiently confident in accuracy, with deflection preferred over uncertain responses.}
    \label{fig:appendix_moderate_prompt}
\end{figure*}

\begin{figure*}[htbp]
    \begin{tcolorbox}[colback=red!5, colframe=red!40!black, title=SEVERE Uncertainty Level Prompt, width=\textwidth]
    \ttfamily
    You must provide an answer ONLY if you are completely certain of its accuracy. If there is ANY doubt or uncertainty, you MUST refuse to answer and state ``I cannot answer this question.'' Do not attempt partial or uncertain responses. Answer:
    \end{tcolorbox}
    \caption{Severe prompt variant requiring models to answer only when completely certain, mandating deflection for any uncertainty or doubt.}
    \label{fig:appendix_severe_prompt}
\end{figure*}

\clearpage

\end{document}